\documentclass[sigconf]{acmart}

\setcopyright{none}
\renewcommand\footnotetextcopyrightpermission[1]{} 
\AtBeginDocument{%
  }
    
\usepackage{graphicx}
\usepackage{colortbl}
\usepackage{array}
\usepackage{multirow}
\usepackage{hhline}
\usepackage{amsmath}

\usepackage{longtable} 
\usepackage{array} 
\usepackage{booktabs} 
\usepackage{ragged2e} 
\newcolumntype{L}[1]{>{\RaggedRight\arraybackslash\hspace{0pt}}p{#1}} 

\usepackage{multirow}
\usepackage{subfig}
\usepackage{float}
\usepackage{multicol}
\usepackage{lipsum}
\usepackage{csquotes}
\usepackage{array}
\usepackage{enumitem}
\usepackage[scientific-notation=true]{siunitx}
\usepackage[normalem]{ulem}
\useunder{\uline}{\ul}{}
\usepackage{array}
\usepackage{algorithmic}
\usepackage[thinlines]{easytable}
\usepackage{graphicx}
\usepackage{textcomp}
\usepackage{xcolor}

\usepackage{colortbl}
\usepackage{multirow}
\usepackage{graphicx}
\usepackage{adjustbox}
\usepackage{array}

\definecolor{lightgreen}{rgb}{0.56, 0.93, 0.56}
\definecolor{lightred}{rgb}{1.0, 0.6, 0.6}
\definecolor{lightblue}{rgb}{0.7, 0.85, 1}

\setcopyright{acmlicensed}
\copyrightyear{2024}
\acmYear{2024}
\acmDOI{XXXXXXX.XXXXXXX}

\acmConference[30th]{ACM SIGKDD Conference on Knowledge Discovery and Data Mining}{ACM KDD August 25 - 29, 2024,
  2024}{Barcelona, Spain}
\acmISBN{978-1-4503-XXXX-X/18/06}

\begin{document}

\title{Agentic Retrieval-Augmented Generation for Time Series Analysis}

\author{Chidaksh Ravuru}
\affiliation{%
  \institution{IIT Dharwad}
  \country{India}
}
\email{200010046@iitdh.ac.in}

\author{Sagar Srinivas Sakhinana}
\affiliation{%
  \institution{TCS Research}
  \country{India}
}
\email{sagar.sakhinana@tcs.com}

\author{Venkataramana Runkana}
\affiliation{%
  \institution{TCS Research}
  \country{India}
}
\email{venkat.runkana@tcs.com}

\vspace{-0mm}
\begin{abstract}
\vspace{0mm}
Time series modeling is crucial for many applications, however, it faces challenges such as complex spatio-temporal dependencies and distribution shifts in learning from historical context to predict task-specific outcomes. To address these challenges, we propose a novel approach using an agentic Retrieval-Augmented Generation (RAG) framework for time series analysis. The framework leverages a hierarchical, multi-agent architecture where the master agent orchestrates specialized sub-agents and delegates the end-user request to the relevant sub-agent. The sub-agents utilize smaller, pre-trained language models (SLMs) customized for specific time series tasks through fine-tuning using instruction tuning and direct preference optimization, and retrieve relevant prompts from a shared repository of prompt pools containing distilled knowledge about historical patterns and trends to improve predictions on new data. Our proposed modular, multi-agent RAG approach offers flexibility and achieves state-of-the-art performance across major time series tasks by tackling complex challenges more effectively than task-specific customized methods across benchmark datasets. 
\vspace{-3mm}
\end{abstract}
 
\maketitle

\vspace{-4mm}
\section{\textbf{Introduction}}
\begin{figure}[ht!]
\centering
\resizebox{0.8\linewidth}{!}{ 
\hspace*{0mm}\includegraphics[keepaspectratio,height=5.0cm,trim=1.25cm 0.10cm 0cm 0.01cm,clip]{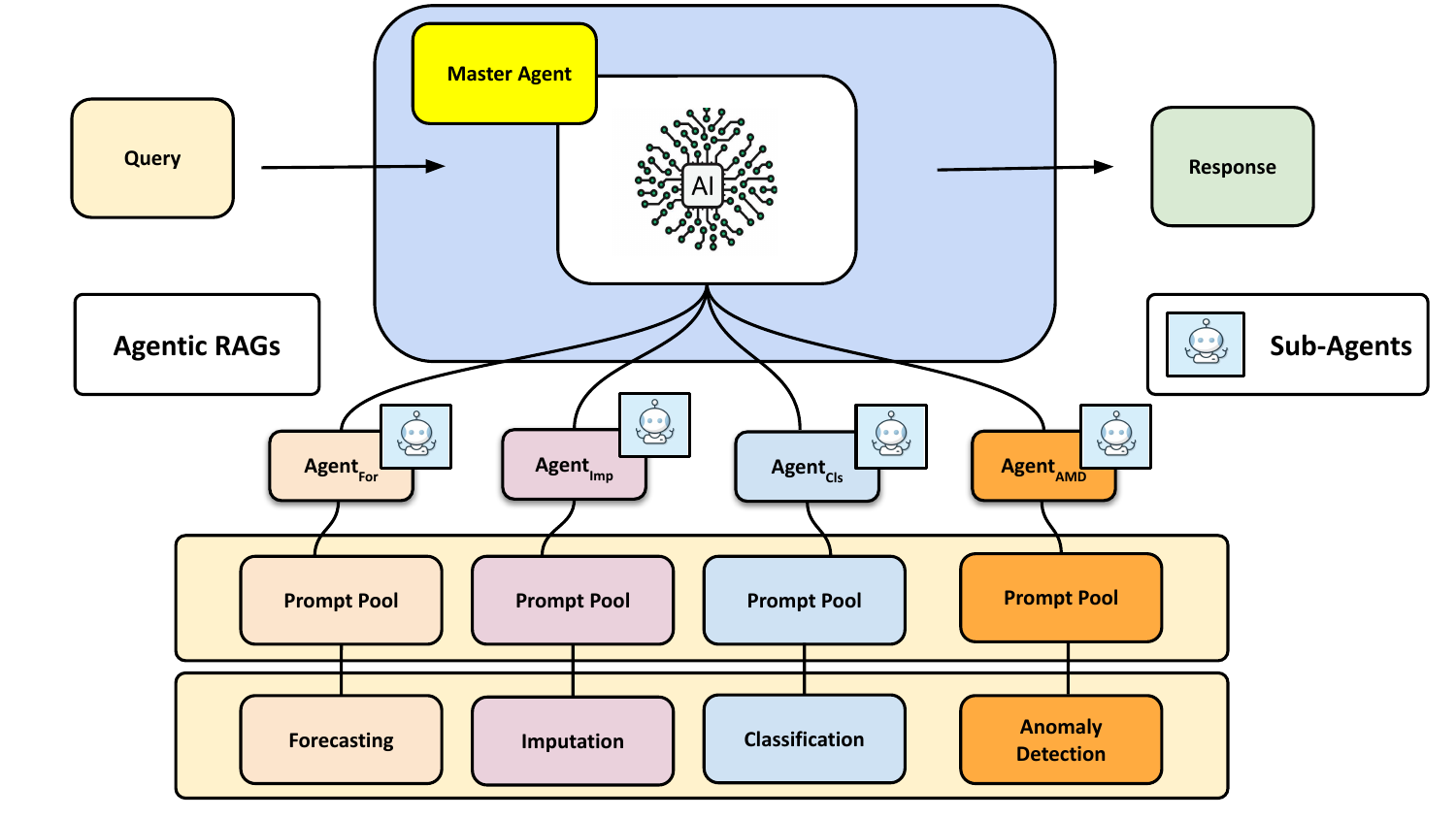} 
}
\vspace{-4mm}
\caption{The figure illustrates the proposed agentic RAG framework, designed to handle diverse time series analysis tasks. The framework employs a hierarchical, multi-agent architecture. A master agent receives end-user questions and routes them to appropriate specialized sub-agents based on the specific time series task (e.g., forecasting, imputation, classification, anomaly detection). The sub-agents utilize pretrained SLMs fine-tuned on task-specific datasets using techniques like instruction tuning and direct preference optimization to capture spatio-temporal dependencies within and across the time series datasets. Each sub-agent maintains its own prompt pool as `key-value' pairs, which stores relevant historical knowledge related to specific trends and patterns within its respective specialized domain. This allows the sub-agents to leverage related past experiences for improved task-specific predictions on new, similar data, and is then relayed back to the user through the master agent. 
}
\label{fig:figure1}
\vspace{-3mm}
\end{figure}

\vspace{-0.5mm}
Time series modeling underpins a vast spectrum of real-world applications, including demand planning \cite{leonard2001promotional}, anomaly detection \cite{zhou2024label}, inventory management \cite{zhou2023business}, energy load forecasting \cite{liu2023sadi}, weather modeling \cite{pathak2022fourcastnet}, and many others. However, it is not without its challenges. High dimensionality, non-linearity, sparsity, and distribution shifts all pose significant hurdles. Successfully navigating these challenges in time series analysis applications necessitates both considerable domain knowledge and the design of neural network architectures tailored to address task-specific goals, leading to better performance. In contrast to task-specific approaches, which employ different architecture designs for time series analysis, foundational pretrained large language models (LLMs), such as OpenAI's GPT-4 \cite{openai2023gpt4} and Google's Gemini \cite{reid2024gemini, team2023gemini}, with their strong generalization and logical reasoning capabilities, have shown remarkable versatility across a broad spectrum of natural language processing (NLP) tasks, requiring minimal fine-tuning\cite{hu2021lora} or only a few demonstrations\cite{brown2020language} for adaptation to niche tasks. Open-source, small-scale pretrained language models (SLMs), such as Google Gemma (\cite{team2024gemma}) and Meta LLaMA (\cite{touvron2023llama, llama3modelcard}), offer cost-effective domain customization through Parameter Efficient Fine-Tuning (PEFT) (\cite{guo2023lq, han2024parameter}) techniques using task-specific labeled datasets. Additionally, these smaller models can be further aligned with human preferences using Direct Preference Optimization (DPO) \cite{christiano2017deep}, a fine-tuning technique that utilizes paired preference data, such as datasets of preferred and dispreferred responses. However, SLMs may lack the reasoning and generalization capabilities of large-scale proprietary language models. The potential of foundational SLMs designed for universal time series applications (a single-model-fits-all approach), such as diverse time series tasks like classification, anomaly detection, forecasting, imputation, and others, remains largely unexplored but holds great promise. This approach contrasts sharply with the traditional approach of using customized, task-specific methods (\cite{zhang2022crossformer, zhang2022grelen, xu2021anomaly}) for time series modeling for various applications. Adapting SLMs designed for NLP tasks for time series modeling to capture trends and patterns within the complex data, though unconventional, offers a clear possibility for providing unique insights. However, this is a challenging task as SLMs are trained primarily on text corpora, which operates on discrete tokens, while time series data is inherently continuous. Furthermore, SLMs may lack the inherent ability to detect and interpret time series patterns and trends like seasonality, cyclicity, or outliers, due to the absence of related pretraining knowledge. Moreover, current LMs designed for time series analysis (\cite{jin2023time, gruver2024large, zhou2024one}) rely on a fixed-length window of past observations to generate predictions, which may be inadequate for capturing complex patterns and trends present in time series data, thus hindering accurate modeling. Smaller window sizes may capture local patterns but miss broader trends, while larger window sizes can capture more context but may overlook finer details. In recent times, Retrieval-Augmented Generation (RAG) or Retrieval-Augmented Language Modeling (RALM)\cite{shi2023replug, ram2023context, lin2023ra} combines pre-trained language models with information retrieval from external knowledge bases to augment text generation capabilities for open-ended question-answering(ODQA)\cite{siriwardhana2023improving} tasks or for improved language modeling for text summarization, completion with improved accuracy. While regular RAG methods augment generation with retrieved knowledge for ODQA tasks, Agentic RAGs take this further by being instruction-following agents that can tackle complex goals through multi-step reasoning and iterative refinement cycles using repeated retrievals over a knowledge base to ensure the final response aligns with the end user request. In this work, we propose an Agentic RAG framework for time series analysis to improve task-specific outcomes by addressing challenges like distributional shifts, fixed window limitations in time series data. Figure ~\ref{fig:figure1} illustrates the framework. Our Agentic RAG framework presents a hierarchical, multi-agent architecture composed of a master (top-level) agent and specialized sub-agents customized for specific time series tasks. The top-level agent acting as the orchestrator analyzes the incoming user request, determines its nature and complexity, and then routes (or delegates) it to the corresponding task-specific sub-agent to produce the desired output. Similarly to how regular RAG frameworks retrieve relevant information from external knowledge bases like documents, databases, or access the real world through APIs, this Agentic RAG framework leverages distinct prompt pools as internal knowledge bases for each sub-agent focused on specific time series tasks. As specialized knowledge repositories tailored to each sub-agent's time series task, the prompt pools store both domain and task-specific knowledge as key-value pairs. This facilitates easy reuse and sharing within and across datasets, promoting knowledge sharing and transfer, reducing the need to relearn or rediscover patterns from scratch. Each `key' represents a specific pattern (seasonality, cyclicality, etc.), and the `value' contains details about that pattern.  When processing new input data, the sub-agent retrieves the most relevant prompts from the pool based on similarity. These prompts provide contextual knowledge about related historical patterns and trends, improving generalization to new scenarios. This knowledge-augmentation approach, by conditioning on past patterns, allows the sub-agent access to a broad spectrum of task-specific knowledge regardless of historical occurrence, enabling it to learn and adapt to diverse trends within complex data for improved predictions. Each sub-agent utilizes pre-trained, SLMs like Gemma\cite{team2024gemma} and Llama 3\cite{llama3modelcard}. We fine-tune each SLM using instruction-tuning on task-specific datasets and optimize them for time series tasks such as forecasting, imputation, or other related tasks. Additionally, we fine-tune using DPO\cite{christiano2017deep} through a dynamic masking technique to align the SLMs task-specific outputs to preferred and non-preferred outcomes, providing adversarial feedback\cite{yoon2019time} through a binary classification task. The master agent for sub-agent orchestration utilizes the 'ReAct' prompting technique\cite{yao2022react}, encouraging the general-purpose SLM to think step-by-step and use external tools (sub-agents, each utilizing a fine-tuned SLM for specific time series tasks) to generate responses. The master agent can even chain sub-agents together to handle complex, multi-step time series analysis tasks, addressing more intricate challenges. However, in this work, the sub-agents operate in isolation, each handling only a single, specific task.

\vspace{-5mm}

In summary, the master agent orchestrates sub-agents, selects the most appropriate sub-agent, and allocates the task to the specialized sub-agent. The sub-agent retrieves relevant information from a shared knowledge base of prompt pools and generates an output based on the retrieved information. The differentiable prompt pools for each sub-agent, acting as specialized dynamic knowledge repositories, provide the necessary historical context and understanding to effectively analyze new input data for their designated tasks. The master agent gathers responses from the chosen sub-agent and synthesize these responses to produce a comprehensive answer for the end-user query. The hierarchical, multi-agent architecture for time series analysis offers key advantages. It enables modularity, flexibility, and accuracy by allowing specialized sub-agents to focus on specific tasks, be updated independently, and be dynamically allocated by the meta-agent to generate comprehensive results. Extensive empirical studies demonstrate that the Agentic-RAG framework achieves performance on par with, or even surpassing, state-of-the-art methods across multiple time series analysis tasks for both univariate and multivariate datasets. The multi-agent approach tackles the diverse and complex challenges of time series analysis, unlike a single, universal agent that attempts to be a jack-of-all-trades for all time series tasks.

\vspace{-4mm}
\section{Problem Formulation}
\vspace{-1mm}
Consider a time series dataset characterized by \(N\) univariate time series, with sequential data collected over \(T\) timestamps, represented as a data matrix \(\mathbf{X} \in \mathbb{R}^{N \times T}\). Each row in this matrix represents a univariate time series, and each column corresponds to data collected at a specific timestamp. To refer to data from a specific time series or timestamp, we use subscripts and superscripts, respectively. For instance, \(X_{i} = \mathbf{X}_{i,:}\) denotes the data from the \(i\)-th time series, and \(X^{t} = \mathbf{X}_{:,t}\) denotes the data at timestamp \(t\).

\vspace{-4mm}
\subsection{Forecasting} 
\vspace{-1mm}
We utilize a sliding window\cite{cini2024taming, yi2024fouriergnn} of size \(\tau\), to construct time series subsequences \(S^{t} = X^{t-\tau+1 : t} \in \mathbb{R}^{N \times \tau}\), which have been observed over previous $\tau$-steps prior to current time step $t$ to  predict about the future values for the next $\nu$-steps, \(S^{t+1} = X^{t+1 : t+\nu} \in \mathbb{R}^{N \times \nu}\).

\vspace{-4mm}
\subsection{Missing Data Imputation} 
\vspace{-1mm}
We utilize a binary mask matrix $\mathbf{M} \in \{0, 1\}^{N \times T}$, where $M_{i, t} = 0$ indicates that the value $X_{i, t}$ is missing, and $M_{i, t} = 1$ indicates that the value is observed in the data matrix $\mathbf{X} \in \mathbb{R}^{N \times T}$. Missing data can follow random or block patterns\cite{marisca2024graph, marisca2022learning, cao2018brits} across the $N$ univariate time series and $T$ timestamps. We utilize observed values $\mathbf{X}_\text{obs} = \mathbf{X} \odot \mathbf{M}$ to estimate the missing values $\mathbf{X}_\text{miss} = \mathbf{X} \odot (1 - \mathbf{M})$. $\odot$ denotes element-wise multiplication. We utilize a sliding window of size $\tau$ over the observed samples $\mathbf{X}_\text{obs}$, to construct subsequences $S^t_\text{obs} = X^{t-\tau+1:t}_\text{obs} \in \mathbb{R}^{N \times \tau}$, which have been observed over previous $\tau$-steps prior to the current time step $t$. These observed samples are used to predict the missing values for the next $\nu$-steps, $S^{t+1}_\text{miss} = X^{t+1:t+\nu}_\text{miss} \in \mathbb{R}^{N \times \nu}$ by leveraging spatio-temporal dependencies within the data.

\vspace{-4mm}
\subsection{Anomaly Detection}
\vspace{-1mm}
Assuming the time series dataset exhibits normal behavior during the initial \(T_{\text{train}}\) timestamps, any pattern deviating from the normal behavior in subsequent timestamps \(t > T_{\text{train}}\) is anomalous. Data observed after \(T_{\text{train}}\) is considered the test dataset. We use a sliding window to construct samples from previous time steps \(S^{t} \in \mathbb{R}^{N \times \tau}\) to predict future values of multiple time series \(S^{t+1} \in \mathbb{R}^{N \times \nu}\). The framework predictions are denoted by $\hat{S}^{t+1} \in \mathbb{R}^{N \times \nu}$. In the unsupervised anomaly detection task, it computes the robust normalized anomaly scores (\resizebox{.0325\textwidth}{!}{$A^{t+1}_{i}$}) for each variable $i$ across the time steps in the training set \resizebox{.035\textwidth}{!}{$\mathcal{T}_{train}$}. This information regarding the variables helps in accurately localizing the anomalies within the test set.

\vspace{0mm}
\resizebox{0.945\linewidth}{!}{
\hspace{0mm}\begin{minipage}{\linewidth}
\begin{equation}
A^{t+1}_{i}=\left|\mathbf{S}^{\mathbf{t+1}}_{i}-\hat{\mathbf{S}}^{t+1}_{i}\right| \nonumber
\end{equation}
\end{minipage}
}

\vspace{1mm}
We compute the simple moving average of the maximum value of anomalousness score(\resizebox{.0325\textwidth}{!}{$A^{t+1}_{i}$}) across the multiple variables at time point $t+1$ over the validation set as given,

\vspace{-2mm}
\resizebox{0.90\linewidth}{!}{
\hspace{0mm}\begin{minipage}{\linewidth}
\begin{gather}
\text{Th} =  \underset{t \in  \mathcal{T}_{val}}{\max}A^{t+1} ; A^{t+1} = \frac{1}{w_{a}} \sum_{t-(w_{a}+1)}^{t+1} \underset{i \in  |N|}{\max} \big(A^{t+1}_{i} \big) 
\end{gather}
\label{displacement}
\end{minipage}
} 

\vspace{1mm}
where $w_{a}$ denotes the number of time points in the moving average calculation. \resizebox{.025\textwidth}{!}{$\mathcal{T}_{val}$} denotes the time points in the validation set. We set the anomaly detection threshold($\text{Th}$) as the moving averaged maximum anomaly value for time $t+1$, \resizebox{.0325\textwidth}{!}{$A^{t+1}$} over the validation data. During inference, time points with an anomaly score above the threshold were flagged as anomalies.

\vspace{-3mm}
\subsection{Classification} 
\vspace{-1mm}
We perform unsupervised $K$-means clustering, identifying ($K$) optimal clusters or regimes and assigning cluster labels $\mathbf{C} \in \mathbb{R}^{T}$ to each time point in the data matrix $\mathbf{X} \in \mathbb{R}^{N \times T}$. Then, a sliding window approach is employed to predict the cluster labels for the next $\nu$   
steps \(S^{t+1} = X^{t+1 : t+\nu} \in \mathbb{R}^{N \times \nu}\) based on the observed sample $S^{t} = X^{t-\tau+1 : t} \in \mathbb{R}^{N \times \tau}$ over the previous $\tau$ time steps.

\vspace{-3mm}
\section{\textbf{Proposed Method}} 
\vspace{0mm}
The proposed framework offers a novel approach to time series analysis by leveraging a hierarchical, multi-agent architecture. It comprises a master agent that coordinates specialized sub-agents, each dedicated to a specific time series task such as forecasting, anomaly detection, or imputation. These sub-agents employ pre-trained language models and utilize prompt pools as internal knowledge bases, storing key-value pairs representing historical patterns and trends. By retrieving relevant prompts from these pools, the sub-agents can augment their predictions with contextual knowledge about related past patterns, enabling them to adapt to diverse trends within complex time series data. The framework's modular design, combined with the strengths of individual sub-agents, allows for improved performance across various time series analysis tasks, surpassing the limitations of traditional fixed-window methods.

\vspace{-3mm}
\subsection{Dynamic Prompting Mechansim}
\vspace{-1mm}
Current time series methods typically utilize past data within a predefined window length to understand historical trends and predict task-specific outcomes. However, this approach may not be optimal because there is no universally ideal window length for all time series data. A larger window length might obscure short-range dependencies, while a smaller window length might fail to capture long-range dependencies . Existing methods fail to capture the full complexity of diverse trends and patterns within the complex data required for accurate time series modeling. Adjusting the window length in real-world scenarios can be challenging and computationally expensive. Achieving this goal is an ambitious task, given the current state of research in this field. To address the challenges of non-stationarity and distributional shifts in real-world data, we utilize a differentiable dynamic prompting mechanism\cite{cao2024tempo}. This mechanism allows traditional time series methods to access related past knowledge by retrieving the same group of prompts from the prompt pool for effective adaptive learning on new, similar input data. The dynamic prompting approach utilizes a shared pool of prompts stored as key-value pairs. For time series applications, each prompt is represented by a key vector encoding the essential global characteristics associated with that prompt. The corresponding value matrix contains specific knowledge related to those trends or patterns, such as seasonality, cyclicality, irregularities, and other effects. The key vector acts as an identifier or query vector to retrieve relevant prompts from the pool based on similarity to the input new data, providing a form of conditioning or context about historical patterns to enhance the predictions. This allows the time series methods to effectively leverage encoded knowledge from past experiences, enhancing their predictions by recognizing and applying learned patterns from the shared prompt pool to the new input data. The pool of prompts $\mathcal{P}$ contains a set of $M$ distinct key-value pairs as follows:

\vspace{-4mm}
\begin{equation}
\mathcal{P} = {(k_1, v_1), (k_2, v_2), \ldots, (k_M, v_M)} \nonumber
\end{equation}

\vspace{-1mm}
Here, $M$ is the total number of prompts in the pool, $k_m \in \mathbb{R}^{\hspace{0.5mm}d}$ is the key vector of the $m$-th prompt, and $v_m \in \mathbb{R}^{\hspace{0.5mm}l \times d}$ is the corresponding prompt value matrix with length $l$ and dimensionality $d$. In order to retrieve the most relevant prompts for a given input time series \( S^{t}_{i} = X^{t-\tau+1 : t}_{i} \in \mathbb{R}^{\tau} \), we first linearly project it into \( d \)-dimensional embeddings \( S^{t}_{i} \in \mathbb{R}^{d} \). We then utilize a score-matching function \(\gamma\) to measure the similarity between the input and each prompt key:

\vspace{-4mm}
\resizebox{0.875\linewidth}{!}{
\begin{minipage}{\linewidth}
\begin{equation}
    \gamma\left(S^{t}_{i}, \boldsymbol{k}_m\right) = \frac{S^{t}_{i} \cdot \boldsymbol{k}_m}{|S^{t}_{i}| |\boldsymbol{k}_m|} \nonumber
\end{equation}
\end{minipage}
}

\vspace{1mm}
where $\gamma$ computes the cosine similarity between the input embedding $S^{t}_{i}$ and the prompt key $\mathbf{k}_m$. The top-$K$ prompts with the highest similarity scores are selected, where $1 \leq K \leq M$. Let $\mathcal{J} = {j_1, j_2, \ldots, j_K}$ be the set of indices corresponding to the top-$K$ most relevant prompts retrieved from the pool $\mathcal{P}$ for the given input time series $S^{t}_{i}$. The selected prompts, along with the original input, are concatenated to form the input embedding $S^{t}_{i}$ as follows:

\vspace{0mm}
\resizebox{0.9\linewidth}{!}{
\begin{minipage}{\linewidth}
\begin{equation}
    S^{t}_{i} = \left[v_{j_{1}} ; \ldots ; v_{j_{K}} ; S^{t}_{i}\right] \nonumber
\end{equation}
\end{minipage}
}

where $\mathbf{s}^{t}_{i} \in \mathbb{R}^{(Kl+1) \times d}$. We linearly project $\mathbf{s}^{t}_{i}$ to $d$-dimensional representation  as follows:

\vspace{-5mm}
\begin{equation}
\hspace{5mm}\mathbf{s}^{t}_{i} = W\mathbf{s}^{t}_{i}  \nonumber
\end{equation}

\vspace{-1mm}
where $W \in \mathbb{R}^{d \times (Kl+1)d}$ is a learnable weight matrix. In summary, it aims to improve time series modeling efficiency  on the task-specific performance by allowing the framework to recognize and apply learned patterns across non-stationarity datasets with distributional shifts via the shared prompt representation pool.

\vspace{-3mm}
\subsection{Fine-Tuning/Preference Optimization SLMs}
Current pretrained SLMs, such as Google's Gemma and Meta's Llama-3 models, are designed with a context length of 8K tokens. However, they struggle to process long input sequences that exceed their pretraining context window. This is because the limited length of the context window during pretraining restricts their effectiveness during inference when dealing with longer texts. SLMs with an improved context length can better capture long-term spatio-temporal dependencies and complex patterns that unfold over extended periods, which is essential for accurate predictions and understanding seasonal or cyclic trends. We build upon recent work \cite{jin2024llm} to improve how SLMs handle long sequences without fine-tuning. A two-tiered attention mechanism (grouped and neighbor attention) allows SLMs to process unseen long-range dependencies, enabling SLMs to naturally handle extended text and maintain performance. It outperforms fine-tuning methods on multiple NLP benchmarks, demonstrating a significant step forward for SLMs in managing long text sequences. Nevertheless, fine-tuning general-purpose SLMs on task-specific data and objectives can still provide significant performance gains and allow for customization and adaptation to the unique challenges and requirements of different time series analysis tasks. Instruction-tuning of SLMs captures complex task-specific spatio-temporal dependencies and improves prediction accuracy. We perform instruction-tuning of SLMs with an improved context length \cite{jin2024llm}(32K tokens) using parameter-efficient fine-tuning (PEFT) techniques on their associated specific tasks (e.g., forecasting, imputation) using the corresponding time-series datasets. This approach could significantly enhance the effectiveness of SLMs in processing extensive time-series data. We leverage Direct Preference Optimization (DPO; \cite{rafailov2024direct}), which involves randomly masking 50 \% of the data and performing binary classification task to predict the corresponding correct task-specific outcomes. This is done to steer the predictions of the SLMs toward more reliable outcomes in the specific context of time series analysis, favoring preferred responses over dispreferred responses.

\vspace{-3mm}  
\section{Experiments}

\vspace{-1mm}
\paragraph{\textbf{Datasets:}} We evaluate the proposed Agentic-RAG framework on four tasks: forecasting, classification, anomaly detection, and imputation. To comprehensively evaluate the framework performance against several baselines, we conducted experiments using both univariate and multivariate benchmark datasets across multiple time series tasks. The variants include Agentic-RAG with SelfExtend-Gemma-2B-instruct, Gemma-7B-instruct, and Llama 3-8B-instruct. We utilized several real-world traffic-related datasets (PeMSD3, PeMSD4, PeMSD7, PeMSD7(M), PeMSD8) obtained from the Caltrans Performance Measurement System (PeMS) \cite{chen2001freeway} for forecasting, classification, and imputation. To ensure consistency with prior research\cite{choi2022graph}, these datasets are preprocessed by aggregating 30-second data points into 5-minute averages. Additionally, publicly available traffic prediction datasets (METR-LA, PEMS-BAY) \cite{Li2018DCRNN} are utilized, with data aggregated into 5-minute intervals, resulting in 288 observations per day. Table~\ref{table:summarydatasets} provides comprehensive details regarding the spatiotemporal multivariate datasets. For anomaly detection, we evaluate the proposed Agentic-RAG framework on publicly available multivariate datasets, conducting a comprehensive benchmark comparison against baseline methods. Table~\ref{table:anomalydatasets} provides an overview of the datasets used in this study. SWaT and WADI\footnote{{https://itrust.sutd.edu.sg/itrust-labs/datasets/}} are real-world datasets on water treatment facilities and distribution networks, respectively. SMAP and MSL are expert annotated open-source datasets of telemetry data sourced from NASA\cite{hundman2018detecting}. The Tennessee Eastman Process (TEP)\footnote{https://dataverse.harvard.edu/dataverse/harvard} dataset is a simulated industrial benchmark designed for process monitoring and control, comprising 20 distinct fault types. The HAI\footnote{https://github.com/icsdataset/hai} dataset comprises time-series data from an industrial testbed for detecting adversarial attacks on industrial control systems, involving steam-turbine power generation and pumped-storage hydropower generation processes, with 38 different attack scenarios. In addition, we discuss the univariate datasets for forecasting and imputation in the technical appendix.

\vspace{-4mm}
\begin{table}[tbhp]
\center
\setlength{\tabcolsep}{0.25em} 
\renewcommand\arraystretch{1.35} 
\centering
 \resizebox{0.45\textwidth}{!}{
\begin{tabular}{c|c|c|c|c|c}
\hline
\textbf{Dataset} & \textbf{Sensors} & \textbf{Timesteps} & \textbf{Time-Range} & \multicolumn{1}{l|}{\textbf{Data Split}} & \multicolumn{1}{l}{\textbf{Granularity}} \\ \hline
PeMSD3 & 358 & 26,208 & 09/2018 - 11/2018 & \multirow{5}{*}{6 / 2 / 2} & \multirow{7}{*}{\rotatebox[origin=c]{270}{5 mins}} \\
PeMSD4 & 307 & 16,992 & 01/2018 - 02/2018 &  &  \\
PeMSD7 & 883 & 28,224 & 05/2017 - 08/2017 &  &  \\
PeMSD8 & 170 & 17,856 & 07/2016 - 08/2016 &  &  \\
PeMSD7(M) & 228 & 12,672 & 05/2012 - 06/2012 &  &  \\ \cline{1-5}
METR-LA & 207 & 34,272 & 03/2012 - 06/2012 & \multirow{2}{*}{7 / 1 / 2} &  \\
PEMS-BAY & 325 & 52,116 & 01/2017 - 05/2017 &  &  \\ \hline
\end{tabular}
}
\vspace{0mm}
\caption{Summary of the spatio-temporal datasets.}
\vspace{-7mm}
\label{table:summarydatasets}
\end{table}

\vspace{-5mm}
\begin{table}[htbp] .
  \centering
  \resizebox{0.75\linewidth}{!}{%
  \begin{tabular}{lrrrrcr}
    \toprule
    Dataset & SWaT & WADI & SMAP & MSL & TEP & HAI\\
    \midrule
    \textbf{Sensors} & 51 & 123 & 25 & 55 & 52 & 59\\
    \textbf{$\tau$} & 25 & 25 & 50 & 55 & 35 & 30\\
   \bottomrule 
\end{tabular}
}
\caption{Statistical summary of benchmark datasets. $\tau$ is the length of subsequences or historical window length.}
\label{table:anomalydatasets}
\vspace{-8mm}
\end{table}

\vspace{-2mm}
\paragraph{\textbf{Evaluation Metrics:}} For forecasting and imputation tasks, the performance of the proposed framework is evaluated using MAE, RMSE, and MAPE metrics on the original scale of the time series data. For classification tasks, we use accuracy. For anomaly detection, we utilize the standard evaluation metrics of precision (P in \%), recall (R in \%), and F1-score (F1 in \%). We utilize a multi-metric approach for a fair and rigorous comparison with baseline models. To do this, we compute the confusion matrix: true positive (TP) for correctly detected anomalies, false negative (FN) for undetected anomalies, true negative (TN) for correctly identified normal points, and false positive (FP) for normal points mistakenly identified as anomalies. Precision (TP/(FP + TP)) represents the proportion of correctly detected anomalies among all identified anomalies, while recall (TP / (FN + TP)) represents the proportion of all true anomalies that were correctly detected. The F1-score is calculated as the harmonic mean of precision and recall. The threshold for identifying anomalies is set to the highest anomaly score(refer to Section \ref{displacement}) from the validation dataset. For the SWaT and WADI datasets, which contain contiguous anomaly segments, we adopt the point adjustment strategy \cite{shen2020timeseries, zhao2020multivariate} to flag the entire subsequence as an anomaly if the model predicts one. On the Tennessee Eastman dataset, we utilize the Fault Detection Rate (FDR, in \%), defined as the ratio of the number of faults detected to the total number of faults that occur, to evaluate the effectiveness of our framework.

\vspace{-2mm}
\paragraph{\textbf{Experimental Settings:}} To reduce memory footprint and computational complexity, we segment the time series datasets using a sliding window technique with a predefined historical window size to obtain time series subsequences (smaller, overlapping sequences of a fixed length). We performed instruction-tuning(fine-tuning) of the small-scale language models, such as SelfExtend-Instruct LLaMA 3-8B, Gemma-2B, and Gemma-7B models using the PEFT technique\cite{xu2023parameter} such as QLoRA\cite{dettmers2024qlora}, on their specific associated time series tasks using corresponding datasets. We set the following hyperparameters: a batch size of 16, a sequence length of 32K, a learning rate of 1e-5, training for 15 epochs, 500 warmup steps, a weight decay of 0.01, and a gradient accumulation of 2 steps. We used the AdamW optimizer\cite{loshchilov2017decoupled} and a linear scheduler to adjust the learning rate during training. We utilized a 4-bit quantization for QLoRA. The QLoRA hyperparameters include the low-rank($r$) of 

\begin{table*}[ht!]
\setlength{\tabcolsep}{0.2em} 
\renewcommand\arraystretch{1.14} 
\centering
 \resizebox{0.87\textwidth}{!}{
\begin{tabular}{c|ccc|ccc|ccc|ccc|ccc}
\hline
\multirow{2}{*}{\textbf{Methods}} & \multicolumn{3}{c|}{\textbf{PeMSD3}} & \multicolumn{3}{c|}{\textbf{PeMSD4}} & \multicolumn{3}{c|}{\textbf{PeMSD7}} & \multicolumn{3}{c|}{\textbf{PeMSD8}} & \multicolumn{3}{c}{\textbf{PeMSD7(M)}} \\ \cline{2-16} 
 & \multicolumn{1}{l}{\textbf{MAE}} & \multicolumn{1}{l}{\textbf{RMSE}} & \multicolumn{1}{l|}{\textbf{MAPE}} & \multicolumn{1}{l}{\textbf{MAE}} & \multicolumn{1}{l}{\textbf{RMSE}} & \multicolumn{1}{l|}{\textbf{MAPE}} & \multicolumn{1}{l}{\textbf{MAE}} & \multicolumn{1}{l}{\textbf{RMSE}} & \multicolumn{1}{l|}{\textbf{MAPE}} & \multicolumn{1}{l}{\textbf{MAE}} & \multicolumn{1}{l}{\textbf{RMSE}} & \multicolumn{1}{l|}{\textbf{MAPE}} & \textbf{MAE} & \textbf{RMSE} & \textbf{MAPE} \\ \hline
HA & 31.58 & 52.39 & 33.78 & 38.03 & 59.24 & 27.88 & 45.12 & 65.64 & 24.51 & 34.86 & 59.24 & 27.88 & 4.59 & 8.63 & 14.35 \\
ARIMA  & 35.41 & 47.59 & 33.78 & 33.73 & 48.80 & 24.18 & 38.17 & 59.27 & 19.46 & 31.09 & 44.32 & 22.73 & 7.27 & 13.20 & 15.38 \\
VAR   & 23.65 & 38.26 & 24.51 & 24.54 & 38.61 & 17.24 & 50.22 & 75.63 & 32.22 & 19.19 & 29.81 & 13.10 & 4.25 & 7.61 & 10.28 \\
FC-LSTM  & 21.33 & 35.11 & 23.33 & 26.77 & 40.65 & 18.23 & 29.98 & 45.94 & 13.20 & 23.09 & 35.17 & 14.99 & 4.16 & 7.51 & 10.10 \\
TCN  & 19.32 & 33.55 & 19.93 & 23.22 & 37.26 & 15.59 & 32.72 & 42.23 & 14.26 & 22.72 & 35.79 & 14.03 & 4.36 & 7.20 & 9.71 \\
TCN(w/o causal)  & 18.87 & 32.24 & 18.63 & 22.81 & 36.87 & 14.31 & 30.53 & 41.02 & 13.88 & 21.42 & 34.03 & 13.09 & 4.43 & 7.53 & 9.44 \\
GRU-ED   & 19.12 & 32.85 & 19.31 & 23.68 & 39.27 & 16.44 & 27.66 & 43.49 & 12.20 & 22.00 & 36.22 & 13.33 & 4.78 & 9.05 & 12.66 \\
DSANet   & 21.29 & 34.55 & 23.21 & 22.79 & 35.77 & 16.03 & 31.36 & 49.11 & 14.43 & 17.14 & 26.96 & 11.32 & 3.52 & 6.98 & 8.78 \\
STGCN  & 17.55 & 30.42 & 17.34 & 21.16 & 34.89 & 13.83 & 25.33 & 39.34 & 11.21 & 17.50 & 27.09 & 11.29 & 3.86 & 6.79 & 10.06 \\
DCRNN  & 17.99 & 30.31 & 18.34 & 21.22 & 33.44 & 14.17 & 25.22 & 38.61 & 11.82 & 16.82 & 26.36 & 10.92 & 3.83 & 7.18 & 9.81 \\
GraphWaveNet  & 19.12 & 32.77 & 18.89 & 24.89 & 39.66 & 17.29 & 26.39 & 41.50 & 11.97 & 18.28 & 30.05 & 12.15 & 3.19 & 6.24 & 8.02 \\
ASTGCN(r)  & 17.34 & 29.56 & 17.21 & 22.93 & 35.22 & 16.56 & 24.01 & 37.87 & 10.73 & 18.25 & 28.06 & 11.64 & 3.14 & 6.18 & 8.12 \\
MSTGCN   & 19.54 & 31.93 & 23.86 & 23.96 & 37.21 & 14.33 & 29.00 & 43.73 & 14.30 & 19.00 & 29.15 & 12.38 & 3.54 & 6.14 & 9.00 \\
STG2Seq    & 19.03 & 29.83 & 21.55 & 25.20 & 38.48 & 18.77 & 32.77 & 47.16 & 20.16 & 20.17 & 30.71 & 17.32 & 3.48 & 6.51 & 8.95 \\
LSGCN  & 17.94 & 29.85 & 16.98 & 21.53 & 33.86 & 13.18 & 27.31 & 41.46 & 11.98 & 17.73 & 26.76 & 11.20 & 3.05 & 5.98 & 7.62 \\
STSGCN  & 17.48 & 29.21 & 16.78 & 21.19 & 33.65 & 13.90 & 24.26 & 39.03 & 10.21 & 17.13 & 26.80 & 10.96 & 3.01 & 5.93 & 7.55 \\
AGCRN  & 15.98 & 28.25 & 15.23 & 19.83 & 32.26 & 12.97 & 22.37 & 36.55 & 9.12 & 15.95 & 25.22 & 10.09 & 2.79 & 5.54 & 7.02 \\
STFGNN  & 16.77 & 28.34 & 16.30 & 20.48 & 32.51 & 16.77 & 23.46 & 36.60 & 9.21 & 16.94 & 26.25 & 10.60 & 2.90 & 5.79 & 7.23 \\
STGODE  & 16.50 & 27.84 & 16.69 & 20.84 & 32.82 & 13.77 & 22.59 & 37.54 & 10.14 & 16.81 & 25.97 & 10.62 & 2.97 & 5.66 & 7.36 \\
Z-GCNETs  & 16.64 & 28.15 & 16.39 & 19.50 & 31.61 & 12.78 & 21.77 & 35.17 & 9.25 & 15.76 & 25.11 & 10.01 & 2.75 & 5.62 & 6.89 \\
STG-NCDE  & 15.57 & 27.09 & 15.06 & 19.21 & 31.09 & 12.76 & 20.53 & 33.84 & 8.80 & 15.45 & 24.81 & 9.92 & 2.68 & 5.39 & 6.76 \\ \hline
\textbf{SelfExtend-Agentic-RAG W/Gemma-2B} & 14.05 & 20.53 & 11.57 & 19.14 & 27.92 & 10.54 & 20.59 & 31.89 & 9.27 & 15.53 & 22.17 & 8.09 & 2.10 & 5.06 & 6.61 \\
\textbf{SelfExtend-Agentic-RAG W/Gemma-7B} & 13.51 & 20.02 & 10.98 & 17.99 & 25.97 & 10.03 & 19.48 & 30.53 & 8.47 & 14.52 & 21.49 & 7.46 & 2.38 & 4.79 & 6.02 \\
\textbf{SelfExtend-Agentic-RAG W/Llama 3 - 8B} & \textbf{13.01} & \textbf{19.48} & \textbf{10.53} & \textbf{17.46} & \textbf{25.54} & \textbf{9.52} & \textbf{19.02} & \textbf{29.97} & \textbf{8.03} & \textbf{14.03} & \textbf{20.98} & \textbf{7.04} & \textbf{2.33} & \textbf{4.68} & \textbf{5.88} \\ \hline
\end{tabular}
}
\vspace{0mm}
\caption{The table compares various methods for 12-sequence-to-12-sequence forecasting tasks on benchmark datasets using multiple evaluation metrics. These methods use 12 past sequences to predict the next 12 sequences.}
\label{table:results1}
\vspace{-5mm}
\end{table*}

\begin{table*}[ht!]
\vspace{-4mm}
\setlength{\tabcolsep}{0.35em} 
\renewcommand\arraystretch{1.1} 
\centering
 \resizebox{0.75\textwidth}{!}{
\begin{tabular}{c|c|ccc|ccc|ccc}
\hline
\multirow{2}{*}{\textbf{Datasets}}  & \multirow{2}{*}{\textbf{Methods}} & \multicolumn{3}{c|}{\textbf{Horizon$\textbf{@}$3}}       & \multicolumn{3}{c|}{\textbf{Horizon$\textbf{@}$6}}       & \multicolumn{3}{c}{\textbf{Horizon$\textbf{@}$12}}       \\ \cline{3-11} 
                                    &                                   & \textbf{RMSE} & \textbf{MAE}  & \textbf{MAPE} & \textbf{RMSE} & \textbf{MAE}  & \textbf{MAPE} & \textbf{RMSE} & \textbf{MAE}  & \textbf{MAPE} \\ \hline
\multirow{13}{*}{\textbf{METR-LA}}  & HA                       & 10.00 & 4.79          & 11.70 & 11.45 & 5.47 & 13.50 & 13.89 & 6.99 & 17.54         \\
                                    & VAR                       & 7.80          & 4.42          & 13.00         & 9.13          & 5.41          & 12.70         & 10.11         & 6.52          & 15.80         \\
                                    & SVR                      & 8.45          & 3.39          & 9.30          & 10.87         & 5.05          & 12.10         & 13.76         & 6.72          & 16.70         \\
                                    & FC-LSTM                   & 6.30          & 3.44          & 9.60          & 7.23          & 3.77          & 10.09         & 8.69          & 4.37          & 14.00         \\
                                    & DCRNN                    & 5.38          & 2.77          & 7.30          & 6.45          & 3.15          & 8.80          & 7.60          & 3.60          & 10.50         \\
                                    & STGCN                     & 5.74          & 2.88          & 7.62          & 7.24          & 3.47          & 9.57          & 9.40          & 4.59          & 12.70         \\
                                    & Graph WaveNet             & 5.15          & 2.69          & 6.90          & 6.22          & 3.07          & 8.37          & 7.37          & 3.53          & 10.01         \\
                                    & ASTGCN                   & 9.27          & 4.86          & 9.21          & 10.61         & 5.43          & 10.13         & 12.52         & 6.51          & 11.64         \\
                                    & STSGCN                    & 7.62          & 3.31          & 8.06          & 9.77          & 4.13          & 10.29         & 11.66         & 5.06          & 12.91         \\
                                    & MTGNN                    & 5.18          & 2.69          & 6.88          & 6.17          & 3.05          & 8.19          & 7.23          & 3.49          & 9.87          \\
                                    & GMAN                     & 5.55          & 2.80          & 7.41          & 6.49          & 3.12          & 8.73          & 7.35          & 3.44          & 10.07         \\
                                    & DGCRN                   & 5.01          & 2.62          & 6.63          & 6.05 & 2.99 & 8.02          & 7.19 & 3.44 & 9.73 \\ \cline{2-11} 
                                    & \textbf{SelfExtend-Agentic-RAG W/Gemma-2B}                & 4.52 & 2.29 & 5.55 & 5.82 & 2.91 & 7.33 & 6.81 & 3.32 & 9.03 \\ 
                                    & \textbf{SelfExtend-Agentic-RAG W/Gemma-7B}              & 4.28 & 2.17 & 5.35 & 5.63 & 2.75 & 7.02 & 6.53 & 3.23 & 8.71 \\ 
                                    & \textbf{SelfExtend-Agentic-RAG W/Llama 3-8B}                & \textbf{4.03} & \textbf{2.02} & \textbf{5.05} & \textbf{5.43} & \textbf{2.61} & \textbf{6.75} & \textbf{6.23} & \textbf{3.12} & \textbf{8.53} \\   \hline \hline
\multirow{13}{*}{\textbf{PEMS-BAY}} & HA                       & 4.30 & 1.89          & 4.16          & 5.82 & 2.50 & 5.62          & 7.54 & 3.31 & 7.65          \\
                                    & VAR                      & 3.16          & 1.74          & 3.60          & 4.25          & 2.32          & 5.00          & 5.44          & 2.93          & 6.50          \\
                                    & SVR                     & 3.59          & 1.85          & 3.80          & 5.18          & 2.48          & 5.50          & 7.08          & 3.28          & 8.01          \\
                                    & FC-LSTM                 & 4.19          & 2.05          & 4.80          & 4.55          & 2.20          & 5.20          & 4.96          & 2.37          & 5.70          \\
                                    & DCRNN                   & 2.95          & 1.38          &  2.90          & 3.97          & 1.74          & 3.90          & 4.74          & 2.07          & 4.90          \\
                                    & STGCN                    & 2.96          & 1.36          & 2.90          & 4.27          & 1.81          & 4.17          & 5.69          & 2.49          & 5.79          \\
                                    & Graph WaveNet          & 2.74          & 1.30          & 2.73          & 3.70          & 1.63          & 3.67          & 4.52          & 1.95          & 4.63          \\
                                    & ASTGCN                   & 3.13          & 1.52          & 3.22          & 4.27          & 2.01          & 4.48          & 5.42          & 2.61          & 6.00          \\
                                    & STSGCN                   & 3.01          & 1.44          & 3.04          & 4.18          & 1.83          & 4.17          & 5.21          & 2.26          & 5.40          \\
                                    & MTGNN                   & 2.79          & 1.32          & 2.77          & 3.74          & 1.65          & 3.69          & 4.49          & 1.94          & 4.53          \\
                                    & GMAN                     & 2.91          & 1.34          & 2.86          & 3.76          & 1.63          & 3.68          & 4.32          & 1.86          & 4.37          \\
                                    & DGCRN                     & 2.69          & 1.28          & 2.66          & 3.63          & 1.59          & 3.55          & 4.42          & 1.89          & 4.43          \\ \cline{2-11} 
                                    & \textbf{SelfExtend-Agentic-RAG W/Gemma-2B}              & 1.81 & 0.91 & 1.82 & 2.71 & 1.31 & 2.71 & 3.31 & 1.72 & 3.32 \\ 
                                    & \textbf{SelfExtend-Agentic-RAG W/Gemma-7B}              & 1.72 & 0.86 & 1.68 & 2.61 & 1.26 & 2.63 & 3.21 & 1.67 & 3.23 \\ 
                                    & \textbf{SelfExtend-Agentic-RAG W/Llama 3-8B}               & \textbf{1.62} & \textbf{0.81} & \textbf{1.63} & \textbf{2.52} & \textbf{1.21} & \textbf{2.51} & \textbf{3.12} & \textbf{1.62} & \textbf{3.14} \\ \hline 
\end{tabular}
}
\vspace{0mm}
\caption{The table compares the performance of various forecasting methods on the METR-LA and PEMS-BAY benchmark datasets using multiple evaluation metrics. All methods use 12 past sequences to predict 3, 6, or 12 future sequences.}
\label{table:results2}
\end{table*}

\begin{table*}[ht!]
\center
\setlength{\tabcolsep}{5pt}
\caption{Experimental results on the anomaly detection benchmark datasets in terms of precision, recall, and F1-score}
\vspace{-4mm}
\label{table:results3}
\resizebox{0.835\linewidth}{!}{%
\begin{tabular}{@{}c|ccc|lll|ccc|ccc|ccc@{}}
\toprule
\multirow{2}{*}{\textbf{Methods}} & \multicolumn{3}{c|}{\textbf{SWaT}}                                  & \multicolumn{3}{c|}{\textbf{WADI}}                                                                         & \multicolumn{3}{c|}{\textbf{SMAP}}                                  & \multicolumn{3}{c}{\textbf{MSL}}    & \multicolumn{3}{c}{\textbf{HAI}}                               \\ \cmidrule(l){2-4} \cmidrule(l){5-7} \cmidrule(l){8-10} \cmidrule(l){11-13} \cmidrule(l){14-16}
                                  & \textbf{P(\%)}       & \textbf{R(\%)}       & \textbf{F1(\%)}          & \multicolumn{1}{c}{\textbf{P(\%)}} & \multicolumn{1}{c}{\textbf{R(\%)}} & \multicolumn{1}{c|}{\textbf{F1}} & \textbf{P(\%)}       & \textbf{R(\%)}       & \textbf{F1(\%)}          & \textbf{P(\%)}       & \textbf{R(\%)}       & \textbf{F1(\%)}     & \textbf{P(\%)}       & \textbf{R(\%)}       & \textbf{F1(\%)}      \\ \midrule
GAN-Li                           & \multicolumn{1}{l}{81.03} & \multicolumn{1}{l}{84.97} & \multicolumn{1}{l|}{77.32} &                                   76.25 &               80.33                     &      77.95                           & 67.10                & 87.06                & 75.19                 & 71.02                & 87.06                & 78.23              &  19.83                 & 18.36                 & 17.45   \\
LSTM-NDT                          & \multicolumn{1}{l}{79.12} & \multicolumn{1}{l}{75.08} & \multicolumn{1}{l|}{78.75} &                                   81.25 &                       78.64             &      75.18                           & 89.65                & 88.46                & 89.05                 & 59.44                & 53.74                & 56.40                &  22.46                 & 23.45                 & 20.32 \\
MTAD-GAT                          & \multicolumn{1}{l}{82.01} & \multicolumn{1}{l}{76.84} & \multicolumn{1}{l|}{72.47} &                                   82.58 &                  84.94                  &      \underline{80.25}                           & 89.06                & 91.23                & 90.41                 & 87.54                & \underline{94.40}                & 90.84       &  24.75                 & 21.78                 & 20.14            \\
MAD-GAN                           & 98.97               & 63.74                & 77.0                 & \multicolumn{1}{c}{41.44}          & \multicolumn{1}{c}{33.92}          & \multicolumn{1}{c|}{37.0}        & 80.49                & 82.14                & 81.31                 & 85.17                & 89.91                & 87.47            &  25.27                 & 23.34                 & 21.87    \\
GDN                               & \underline{99.35}                & 68.12                & 81.0                 & \multicolumn{1}{c}{\underline{97.50}}          & \multicolumn{1}{c}{40.19}          & \multicolumn{1}{c|}{57.0}        & 86.62                & 84.27                & 83.24                 & 89.92               & 87.24                & 86.84           &  43.41                 & 46.27                 & \underline{44.59}     \\
GTA                         & 74.91                &  \underline{96.41}               & 84.0                 & \multicolumn{1}{c}{74.56}          & \multicolumn{1}{c}{\underline{90.50}}          & \multicolumn{1}{c|}{82.0}        & 89.11                & 91.76                & 90.41                 & 91.04                & 91.17                & 91.11          &  44.91                 & 41.63                 & 40.29        \\ 
LOF                               & 72.15                & 65.43                & 68.62                &      57.02                              &                                   61.17 &     53.46                            & 58.93                & 56.33                & 57.60                 & 47.72                & 85.25                & 61.18       &  31.27                 & 29.93                 & 26.48\\
Deep-SVDD                        & 80.42                & 84.45                & 82.39                &                74.18                    &                                   70.82 &              73.43                   & 89.93                & 56.02                & 69.04                 & 91.92                & 76.63                & 83.58                 &  34.81                 & 31.26                 & 30.94\\
DAGMM                            & 89.92                & 57.84                & 70.4                  & \multicolumn{1}{c}{54.44}          & \multicolumn{1}{c}{26.99}          & \multicolumn{1}{c|}{36.0}        & 86.45                & 56.73                & 68.51                 & 89.60                & 63.93                & 74.62          &  35.56                 & 37.12                 & 33.77       \\  
MMPCACD                           & 82.52                & 68.29                & 74.73                &         74.29                           &                                   75.01 &                     71.48            & 88.61                & 75.84                & 81.73                 & 81.42                & 61.31                & 69.95                 &  31.58                 & 29.46                 & 27.33 \\
VAR                               & 81.59                & 60.29                & 69.34                &       75.59                             &                                   69.36 &     66.21                            & 81.38                & 53.88                & 64.83                 & 74.68                & 81.42                & 77.90 &  34.42                 & 36.28                 & 31.97\\
LSTM                              & 86.15                & 83.27                & 84.69                &    68.73                                &                 62.47                   &       65.74                          & 89.41                & 78.13                & 83.39                 & 85.45                & 82.50                & 83.95              &  35.61                 & 32.84                 & 31.92   \\
CL-MPPCA                          & 76.78                & 81.50                & 79.07                &   69.72                                 &    65.23                                &     67.32                            & 86.13                & 63.16                & 72.88                 & 73.71                & 88.54                & 80.44                 &  33.82                 & 31.74                 & 30.05 \\
ITAD                              & 63.13                & 52.08                & 57.08                &     71.95                               &                                   69.39 &      65.76                          & 82.42                & 66.89                & 73.85                 & 69.44                & 84.09                & 76.07                 &  36.72                 & 33.42               & 32.47\\
LSTM-VAE                           & 76.00                & 89.50                & 82.20                 & \multicolumn{1}{c}{87.79}          & \multicolumn{1}{c}{14.45}          & \multicolumn{1}{c|}{25.0}        & 92.20                & 67.75                & 78.10                 & 85.49                & 79.94                & 82.62       &  38.25                 & 37.94               & 35.04         \\
BeatGAN                           & 64.01                & 87.46                & 73.92                &    74.46                                &                                   70.71 &     76.52                            & 92.38                & 55.85                & 69.61                 & 89.75                & 85.42                & 87.53 &  39.41                 & 38.03               & 35.47\\
OmniAnomaly                      & 81.42                & 84.30                & 82.83                &        78.18                            &                                   80.13 &         77.24                        & 92.49                & 81.99                & 86.92                 & 89.02                & 86.37                & 87.67 &  46.29                 & 43.75               & 42.73\\
InterFusion                       & 80.59                & 85.58                & 83.01                &         81.78                           &           84.37                         &    80.21                             & 89.77                & 88.52                & 89.14                 & 81.28                & 92.70                & 86.62  &  45.72                 & 43.15               & 42.55\\
THOC                              & 83.94                & 86.36                & 85.13                &    84.24                                 &    81.32                                 &     80.09                             & 92.06                & 89.34                & 90.68                 & 88.45                & 90.97                & 89.69               &  43.72                 & \underline{45.82}               & 43.67  \\
GRELEN                                & 95.60                & 83.50                & \underline{89.10}                &      77.30                              &                                   61.30 &        68.20                         & \underline{94.45}                & \underline{98.16}                & \underline{97.29}                 & \underline{94.36}                & 94.04                & \underline{91.58}       &  \underline{47.31}                 & 43.12               & 40.58          \\ 
\textbf{Agentic-RAG W/Gemma-2B}                          & \multicolumn{1}{l}{99.35} & \multicolumn{1}{l}{98.00} & \multicolumn{1}{l|}{92.45} & \multicolumn{1}{l}{98.50} & \multicolumn{1}{l}{91.85} & \multicolumn{1}{l|}{89.95} & \multicolumn{1}{l}{98.10} & \multicolumn{1}{l}{98.85} & \multicolumn{1}{l|}{98.90} & \multicolumn{1}{l}{97.95} & \multicolumn{1}{l}{97.25} & \multicolumn{1}{l}{96.90} & \multicolumn{1}{l}{58.10} & \multicolumn{1}{l}{56.00} & \multicolumn{1}{l}{53.10} \\ 
\textbf{Agentic-RAG W/Gemma-7B}                          & \multicolumn{1}{l}{99.42} & \multicolumn{1}{l}{98.08} & \multicolumn{1}{l|}{92.53} & \multicolumn{1}{l}{98.58} & \multicolumn{1}{l}{91.93} & \multicolumn{1}{l|}{90.03} & \multicolumn{1}{l}{98.18} & \multicolumn{1}{l}{98.93} & \multicolumn{1}{l|}{98.98} & \multicolumn{1}{l}{98.03} & \multicolumn{1}{l}{97.33} & \multicolumn{1}{l}{96.98} & \multicolumn{1}{l}{58.18} & \multicolumn{1}{l}{56.08} & \multicolumn{1}{l}{53.18} \\ 
\textbf{Agentic-RAG W/Llama-8B}                          & \multicolumn{1}{l}{99.47} & \multicolumn{1}{l}{98.15} & \multicolumn{1}{l|}{92.59} & \multicolumn{1}{l}{98.63} & \multicolumn{1}{l}{91.97} & \multicolumn{1}{l|}{90.08} & \multicolumn{1}{l}{98.24} & \multicolumn{1}{l}{98.97} & \multicolumn{1}{l|}{99.04} & \multicolumn{1}{l}{98.11} & \multicolumn{1}{l}{97.37} & \multicolumn{1}{l}{97.04} & \multicolumn{1}{l}{58.27} & \multicolumn{1}{l}{56.13} & \multicolumn{1}{l}{53.24} \\  \bottomrule
\multicolumn{10}{l}{Best performance in bold. Second-best with underlines(except Agentic-RAG framework Variants).} \\
\end{tabular}
}
\vspace{-4mm}
\end{table*}

\begin{table*}[htbp]
\centering
\setlength{\tabcolsep}{2.5pt}
\caption{Experimental results on simulated Tennessee Eastman dataset in terms of fault detection rate (FDR(\%))}
\vspace{-4mm}
\label{table:results4}
\resizebox{0.94\linewidth}{!}{%
\begin{tabular}{@{}c|cccccccccccccccccccc|c@{}}
\toprule
\textbf{Base Model} & \textbf{1} & \textbf{2} & \textbf{3} & \textbf{4} & \textbf{5} & \textbf{6} & \textbf{7} & \textbf{8} & \textbf{9} & \textbf{10} & \textbf{11} & \textbf{12} & \textbf{13} & \textbf{14} & \textbf{15} & \textbf{16} & \textbf{17} & \textbf{18} & \textbf{19} & \textbf{20}  \\ \midrule
Transformer & 99.64 & 98.45 & 5.00 & 99.96 & 28.86 & 100 & 100 & 96.43 & 5.19 & 17.48 & 77.51 & 98.20 & 94.01 & 99.97 & 5.39 & 13.43 & 91.53 & 93.76 & 25.13 & 48.05 \\
TCN & 99.61 & 97.93 & 5.12 & 100 & 26.46 & 100 & 100 & 94.68 & 5.19 & 35.57 & 80.51 & 96.63 & 93.48 & 99.97 & 5.36 & \underline{21.10} & 96.14 & 93.90 & 23.39 & 47.92 \\
FNet & 99.67 & 98.64 & 4.86 & 99.18 & 25.82 & 100 & 100 & 96.76 & 18.87 & 18.87 & 76.08 & 98.11 & 94.07 & 99.96 & 5.48 & 13.74 & 91.05 & 93.70 & 24.43 & 45.59 \\
GTA & 98.12 & \underline{99.35} & 5.88 & 98.04 & \underline{55.82} & 100 & 100 & 97.34 & 20.18 & 34.33 & 79.81 & 98.72 & 96.03 & 98.21 & 7.64 & 16.69 & 92.25 & 94.78 & 26.57 & 47.31 \\
GDN & \textbf{99.81} & 99.27 & 6.72 & 99.56 & 41.07 & 100 & 100 & 95.04 & 16.46 & 41.22 & 79.57 & \underline{99.64} & 95.71 & 97.58 & 7.83 & 15.64 & 92.79 & \underline{95.27} & 27.17 & 48.81 \\
MTAD-GAT & \underline{99.78} & 98.91 & 8.92 & 99.81 & 39.33 & 100 & 100 & \underline{98.57} & \underline{20.37} & 43.93 & \underline{82.47} & 99.51 & \underline{96.84} & 99.74 & \underline{10.13} & 16.98 & 94.47 & 94.60 & \underline{30.79} & 58.90 \\
GRELEN & 99.67 & 98.64 & \underline{10.86} & 99.18 & 51.82 & 100 & 100 & 96.76 & 18.87 & \underline{48.87} & 76.08 & 98.11 & 94.07 & 99.96 & 5.48 & 13.74 & 91.05 & 93.70 & 24.43 & 62.59 \\
\textbf{Agentic-RAG W/Gemma-2B} & \textbf{99.60} & \textbf{99.75} & \textbf{16.10} & \textbf{99.85} & \textbf{75.20} & \textbf{99.85} & \textbf{99.85} & \textbf{99.30} & 28.90 & \textbf{68.00} & \textbf{87.00} & \textbf{99.30} & \textbf{98.50} & \textbf{99.60} & \textbf{13.80} & 29.20 & \textbf{99.70} & \textbf{98.05} & 41.10 & 79.20 \\ 
\textbf{Agentic-RAG W/Gemma-7B} & \textbf{99.66} & \textbf{99.82} & \textbf{16.18} & \textbf{99.90} & \textbf{75.28} & \textbf{99.90} & \textbf{99.90} & \textbf{99.40} & 29.00 & \textbf{68.12} & \textbf{87.10} & \textbf{99.35} & \textbf{98.58} & \textbf{99.68} & \textbf{13.88} & 29.30 & \textbf{99.78} & \textbf{98.13} & 41.18 & 79.28 \\ 
\textbf{Agentic-RAG W/Llama-8B} & \textbf{99.72} & \textbf{99.89} & \textbf{16.23} & \textbf{100} & \textbf{75.38} & \textbf{100} & \textbf{100} & \textbf{99.47} & \textbf{29.04} & \textbf{68.16} & \textbf{87.15} & \textbf{99.46} & \textbf{98.64} & \textbf{99.75} & \textbf{13.96} & 29.37 & \textbf{99.83} & \textbf{98.21} & 41.23 & 79.35 \\ 
\bottomrule
\multicolumn{10}{l}{Best performance in bold. Second-best with underlines(except Agentic-RAG framework Variants).} \\
\end{tabular}
}
\vspace{-5mm}
\end{table*}

16, an $\alpha$ of 32, and a dropout of 0.05 to ensure efficient parameter updates. We performed preference tuning on the SLMs using Direct Preference Optimization(DPO\cite{rafailov2024direct}) along with QLoRA, minimizing the binary cross-entropy (BCE) loss with the following hyperparameters: a learning rate of 5.0e-7 with a cosine scheduler and a gradient accumulation of 2 steps. $\beta$ was set to 0.2 to better align SLMs with the desired preferences. We conducted training for 3 epochs using the AdamW optimizer, with a batch size of 8 for both the training and evaluation phases. These hyperparameters were chosen to balance the trade-off between SLMs' performance on the specific time series task and computational resources. Optimal hyperparameter values are highly task-specific and depend on the dataset and language model architecture. Extensive experimentation are crucial to find the best configurations. We discuss the hyperparameter optimization results in appendix. To ensure efficient and consistent framework training, we preprocess time-series data by standardizing each variable (zero mean, unit variance) and calculate evalution metric on the original scale. We leverage NVIDIA GPUs and PyTorch for accelerated training, enabling the use of small-scale models and datasets. For robust evaluation, we conduct multiple independent runs and report ensemble averages.

\vspace{-2mm}
\section{Results}
\vspace{-1mm}
Tables \ref{table:results1}-\ref{table:results2} present a performance comparison of the Agentic-RAG framework variants with baseline methods on seven benchmark datasets (PeMSD3, PeMSD4, PeMSD7, PeMSD7M, PeMSD8, METR-LA, and PEMS-BAY) on the forecasting task. We report experimental results from a previous study \cite{choi2022graph} for a fair and rigorous comparison. Tables \ref{table:results3}-\ref{table:results4} show the performance of Agentic-RAG framework variants on time-series anomaly detection on benchmark datasets. We present experimental results of baseline methods from earlier studies \cite{xu2021anomaly, deng2021graph, chen2021learning, fu2022mad}. Our proposed framework outperforms baseline methods across the benchmark datasets, showing significant improvements on the forecasting and anomaly detection tasks. We present experimental results on missing data imputation and classification tasks in the appendix. Experimental results on univariate datasets across all time series tasks are discussed in the appendix.

\vspace{-3mm}
\section{Conclusion}
\vspace{-1mm}
In this work, we propose an Agentic RAG framework to address the challenges of distribution shifts, and fixed-length subsequences in time series analysis. The framework overcomes these challenges by leveraging a hierarchical, multi-agent architecture with specialized sub-agents for various time series tasks. Each sub-agent utilizes a prompt pool as its internal knowledge base to store historical patterns and trends. The sub-agent retrieves relevant prompts and utilizes the corresponding knowledge to improve predictions on new, unseen data. This modular design with task-specific sub-agents and knowledge augmentation outperforms traditional methods in handling complex time series analysis tasks.

\clearpage
\newpage

\bibliographystyle{ACM-Reference-Format}
\bibliography{sample-base}

\newpage
\clearpage
\appendix

\section{Multivariate Spatio-Temporal Datasets}

\vspace{1mm}
\subsection{\textbf{Missing Data Imputation}} 
\vspace{1mm}
Time series imputation is a critical step in time series analysis. It addresses a common issue in this field: missing values within datasets. These missing values can arise from sensor failures, data transmission errors, or incomplete records. By imputing these gaps, time series imputation ensures the quality and reliability of subsequent analyses. The Agentic-RAG framework achieves this by handling seasonality, trends and capturing the inherent spatio-temporal dependencies within the data. Ultimately, imputation improves data quality, enabling more accurate analysis, modeling, and decision-making. In essence, it plays a vital role by maintaining data integrity and enabling reliable analysis. To evaluate the Agentic-RAG framework's ability to handle missing data, we simulated two types of missingness patterns: point missing and block missing\cite{roth2022forecasting, cini2021multivariate}. These patterns represent varying degrees of data availability. To achieve this, we introduced synthetic missingness into time series datasets following these patterns. For point missing, individual values were randomly omitted with a probability threshold ($p$), controlling the overall percentage of missing data. The block missing pattern involves removing contiguous, multi-period, multi-time series segments. This is done by randomly selecting start and end times, as well as start and end time series, to define uniform blocks with an average length of (\l). All data points within each block are then omitted. Furthermore, two block missing patterns are considered: temporal and spatial. For temporal block missing, contiguous multi-period segments are removed from a given time series. This is done by randomly selecting start and end times, creating stretches of unavailable temporal data. For spatial block missing, contiguous blocks are removed across multiple related time series at specific time points. This involves randomly selecting the start and end time series, resulting in missing spatial data at the chosen time points. Both patterns show varying levels of missing information in the time series data. In summary, point missing refers to sporadic gaps in the data, while block missing involves the absence of entire contiguous multi-period and multi-series segments. Block missing can further be categorized into two types: temporal block missing, where contiguous segments are removed within a single time series, and spatial block missing, where contiguous blocks are removed across multiple related time series, mimicking realistic scenarios of faulty data collection. In the context of time series imputation, ``in-sample" and ``out-of-sample" imputation refer to distinct evaluation settings. In-sample imputation involves the imputation method reconstructing missing values within a given fixed input sequence, $S^{t}$, using all available observed data within that sequence. Out-of-sample imputation involves training the imputation method using the fixed sequence $S^{t}$ to impute missing points in a future sequence, $S^{t+1}$. In this work, we utilize out-of-sample settings, as this approach mimics real-world scenarios and rigorously assesses the Agentic-RAG framework's robustness and generalizability by evaluating its ability to handle new, unseen data. The simulated datasets with missing values were then used to evaluate the missing data handling capabilities of the proposed Agentic-RAG framework. We split multiple benchmark datasets in chronological order with a ratio of 7:1:2 for the METR-LA and PEMS-BAY datasets and a ratio of 6:2:2 for the other datasets into training, validation, and test sets. We evaluated the Agentic-RAG framework's performance on simulated data using multiple imputation metrics (e.g., RMSE, MAE, and MAPE). This analysis helps us understand how well the framework handles time series data with missing values, particularly how its performance changes as the percentage of missing data increases.  We establish the Agentic-RAG framework, trained on complete data (no missing values), as a strong performance benchmark. This benchmark allows us to evaluate the framework's effectiveness in imputing missing data under different conditions of data incompleteness. Tables \ref{table:results5} and \ref{table:results6} present the imputation results on standard benchmark datasets with different missingness patterns, while the framework performs slightly worse than the baseline for minimal missing data. Its accuracy degrades more significantly as the data becomes more incomplete, regardless of the specific missingness pattern. Our proposed Agentic-RAG framework demonstrates robustness to missing data by focusing on the available observations for imputing missing values, thereby avoiding the introduction of potentially inaccurate estimates that could obscure the underlying trends and patterns within the time series data. Additionally, the Agentic-RAG framework effectively captures the complex non-linear intra- and inter-time series dependencies and this leads to more reliable imputation. The experiments show that our framework can learn the spatiotemporal dependencies from partially observed data with various missingness patterns, resulting in lower imputation errors.

\vspace{-3mm}
\subsection{\textbf{Time Series Classification}} 
Time series classification is a crucial task with applications across various domains. In time series analysis, regimes, or clusters represent distinct behavioral modes, operating conditions, or states of the system underlying the data. Identifying and characterizing these regimes is crucial for understanding the complex patterns and dynamics within the data. This allows for more accurate modeling, forecasting, and decision-making in applications where time series analysis is essential. The emergence of different regimes or clusters can stem from changes in the data generation process, external conditions, or the inherent non-stationarity and multivariate nature of the time series. This reflects the rich information content and complexity often encountered in real-world time series data. To evaluate the proposed Agentic-RAG framework's ability to handle time series classification tasks, an unsupervised clustering approach was employed for data labeling. We first applied k-means clustering to the original time series datasets, determining the optimal number of clusters (k) using established techniques such as the elbow method or silhouette analysis. The optimal clusters were treated as class labels, representing distinct regimes within the time series, and each time series was assigned the corresponding cluster label, creating a labeled classification dataset. We adopted a time-based division strategy to split multiple benchmark datasets into training, validation, and testing sets. The METR-LA and PEMS-BAY datasets were split at a 7:1:2 ratio, while other datasets used a 6:2:2 split. We evaluated the framework's performance on the held-out test set using standard classification metrics: accuracy, precision, recall. This methodology allowed us to assess the framework's ability to learn the underlying patterns and relationships associated with

\begin{table*}[ht!]
\setlength{\tabcolsep}{0.3em} 
\renewcommand\arraystretch{1.3} 
\centering
\resizebox{0.925\textwidth}{!}{
\begin{tabular}{c|c|ccc|ccc|ccc|ccc}
\hline
\multirow{2}{*}{\textbf{Missing Scheme}} & \multirow{2}{*}{\textbf{Missing Rate}} & \multicolumn{3}{c|}{\textbf{PeMSD3}} & \multicolumn{3}{c|}{\textbf{PeMSD4}} & \multicolumn{3}{c|}{\textbf{PeMSD7}} & \multicolumn{3}{c}{\textbf{METR-LA}} \\ \cline{3-14} 
 &  & \textbf{RMSE} & \textbf{MAE} & \textbf{MAPE} & \textbf{RMSE} & \textbf{MAE} & \textbf{MAPE} & \textbf{RMSE} & \textbf{MAE} & \textbf{MAPE} & \textbf{RMSE} & \textbf{MAE} & \textbf{MAPE} \\ \hline
\textbf{SelfExtend-Agentic-RAG w/Llama-8B} & \textbf{0\%} & \textbf{19.48} & \textbf{13.01} & \textbf{10.53} & \textbf{25.54} & \textbf{17.46} & \textbf{9.52} & \textbf{29.97} & \textbf{19.02} & \textbf{8.03} & \textbf{6.23} & \textbf{3.12} & \textbf{8.53} \\ \hline
\multirow{3}{*}{Point} & 10\% & 21.12 & 14.07 & 12.15 & 28.23 & 19.18 & 11.04 & 32.11 & 20.06 & 10.12 & 7.05 & 4.01 & 10.13 \\
 & 30\% & 22.55 & 15.23 & 13.32 & 30.61 & 20.62 & 12.63 & 34.62 & 21.58 & 11.64 & 7.82 & 4.51 & 11.02 \\
 & 50\% & 24.14 & 16.39 & 14.29 & 33.17 & 22.21 & 14.08 & 37.24 & 23.15 & 13.21 & 8.57 & 5.03 & 12.18 \\ \hline
\multirow{3}{*}{Block} & 10\% & 25.07 & 17.14 & 15.25 & 35.18 & 23.14 & 15.18 & 39.21 & 25.19 & 14.13 & 9.04 & 5.53 & 13.12 \\
 & 30\% & 27.21 & 18.45 & 16.48 & 38.28 & 25.12 & 17.23 & 42.32 & 27.07 & 16.27 & 10.09 & 6.02 & 14.57 \\
 & 50\% & 29.18 & 20.09 & 18.19 & 41.23 & 27.11 & 19.16 & 45.27 & 29.03 & 18.12 & 11.11 & 6.53 & 16.07 \\ \hline 
\multirow{3}{*}{\begin{tabular}[c]{@{}c@{}}Block\\ (Only Spatial)\end{tabular}} & 10\% & 23.04 & 15.59 & 13.42 & 31.19 & 21.23 & 13.09 & 35.18 & 22.14 & 12.61 & 8.02 & 4.53 & 11.59 \\
 & 30\% & 25.09 & 17.23 & 15.18 & 34.26 & 23.15 & 15.12 & 38.25 & 24.19 & 14.21 & 9.11 & 5.02 & 13.13 \\
 & 50\% & 27.15 & 18.52 & 16.59 & 37.23 & 25.18 & 17.19 & 41.16 & 26.13 & 16.17 & 10.14 & 5.57 & 14.52 \\ \hline
 \multirow{3}{*}{\begin{tabular}[c]{@{}c@{}}Block\\ (Only Temporal)\end{tabular}} & 10\% & 22.57 & 15.12 & 13.18 & 30.62 & 20.53 & 13.07 & 34.53 & 21.48 & 11.64 & 7.81 & 4.52 & 11.19 \\
 & 30\% & 24.62 & 16.48 & 14.53 & 33.72 & 22.48 & 15.27 & 37.58 & 23.41 & 13.58 & 8.89 & 5.08 & 12.59 \\
 & 50\% & 26.48 & 18.19 & 16.32 & 36.53 & 24.31 & 18.02 & 40.42 & 25.38 & 15.43 & 9.76 & 5.53 & 14.07 \\ \hline
\end{tabular}
}
\vspace{1mm}
\caption{The table presents the Agentic-RAG framework's evaluation results on various metrics for missing data imputation across PeMSD3, PeMSD4, PeMSD7, and METR-LA benchmark datasets with diverse missing data patterns.}
\label{table:results5}
\vspace{-5mm}
\end{table*}

\vspace{-5mm}
\begin{table*}[ht!]
\setlength{\tabcolsep}{0.45em} 
\renewcommand\arraystretch{1.05} 
\centering
\resizebox{0.9\textwidth}{!}{
\begin{tabular}{c|c|ccc|ccc|ccc}
\hline
\multirow{2}{*}{\textbf{Missing Scheme}} & \multirow{2}{*}{\textbf{Missing Rate}} & \multicolumn{3}{c|}{\textbf{PeMSD7(M)}} & \multicolumn{3}{c|}{\textbf{PeMSD8}} & \multicolumn{3}{c}{\textbf{PEMS-BAY}} \\ \cline{3-11} 
 &  & \textbf{MAE} & \textbf{RMSE} & \textbf{MAPE} & \textbf{MAE} & \textbf{RMSE} & \textbf{MAPE} & \textbf{MAE} & \textbf{RMSE} & \textbf{MAPE} \\ \hline
\textbf{SelfExtend-Agentic-RAG w/Llama-8B} & \textbf{0\%} & \textbf{2.33} & \textbf{4.68} & \textbf{5.88} & \textbf{14.03} & \textbf{20.98} & \textbf{7.04} & \textbf{1.62} & \textbf{3.12} & \textbf{3.14} \\ \hline
\multirow{3}{*}{Point} & 10\% & 2.46 & 4.75 & 6.12 & 15.14 & 22.12 & 7.58 & 1.72 & 3.26 & 3.28 \\
 & 30\% & 2.68 & 5.02 & 6.43 & 16.27 & 23.18 & 8.12 & 1.83 & 3.41 & 3.42 \\
 & 50\% & 2.89 & 5.27 & 6.73 & 17.32 & 24.29 & 8.69 & 1.94 & 3.56 & 3.57 \\ \hline
\multirow{3}{*}{Block} & 10\% & 2.61 & 4.89 & 6.37 & 15.75 & 22.98 & 7.89 & 1.79 & 3.34 & 3.34 \\
 & 30\% & 2.84 & 5.21 & 6.68 & 16.92 & 23.99 & 8.42 & 1.89 & 3.48 & 3.48 \\
 & 50\% & 3.07 & 5.53 & 7.03 & 18.12 & 25.08 & 8.98 & 2.01 & 3.63 & 3.63 \\ \hline
\multirow{3}{*}{\begin{tabular}[c]{@{}c@{}}Block\\ (Spatial Only)\end{tabular}} & 10\% & 2.55 & 4.81 & 6.23 & 15.49 & 22.68 & 7.75 & 1.75 & 3.31 & 3.31 \\
 & 30\% & 2.78 & 5.12 & 6.56 & 16.67 & 23.74 & 8.28 & 1.86 & 3.46 & 3.46 \\
 & 50\% & 3.00 & 5.41 & 6.88 & 17.89 & 24.89 & 8.83 & 1.97 & 3.60 & 3.60 \\ \hline 
\multirow{3}{*}{\begin{tabular}[c]{@{}c@{}}Block\\ (Temporal Only)\end{tabular}} & 10\% & 2.52 & 4.78 & 6.18 & 15.37 & 22.58 & 7.72 & 1.74 & 3.29 & 3.29 \\
 & 30\% & 2.75 & 5.09 & 6.51 & 16.52 & 23.62 & 8.24 & 1.85 & 3.44 & 3.44 \\
 & 50\% & 2.98 & 5.38 & 6.83 & 17.75 & 24.76 & 8.80 & 1.96 & 3.58 & 3.58 \\ \hline
\end{tabular}
}
\vspace{1mm}
\caption{The table presents the performance of the Agentic-RAG framework in imputing missing data on the PeMSD7(M), PeMSD8, and PEMS-BAY benchmark datasets with the various synthetic missing data patterns.}
\label{table:results6}
\vspace{-5mm}
\end{table*}

each cluster/class and its overall effectiveness in classifying time series data based on inherent complex spatio-temporal regimes, paving the way for its practical application in real-world scenarios. The experimental results, presented in Tables \ref{table:results7} and \ref{table:results8}, show a comparison with the simple baselines. 

\begin{table*}[ht!]
\setlength{\tabcolsep}{0.3em} 
\renewcommand\arraystretch{1.35} 
\centering
\resizebox{1.00\textwidth}{!}{
\begin{tabular}{c|ccc|ccc|ccc|ccc}
\hline
\multirow{2}{*}{\textbf{Dataset}} & \multicolumn{3}{c|}{\textbf{PeMSD3}} & \multicolumn{3}{c|}{\textbf{PeMSD4}} & \multicolumn{3}{c|}{\textbf{PeMSD7}} & \multicolumn{3}{c}{\textbf{METR-LA}} \\ \cline{2-13} 
 & \textbf{Accuracy} & \textbf{Precision} & \textbf{Recall} & \textbf{Accuracy} & \textbf{Precision} & \textbf{Recall} & \textbf{Accuracy} & \textbf{Precision} & \textbf{Recall} & \textbf{Accuracy} & \textbf{Precision} & \textbf{Recall} \\ \hline
\textbf{SelfExtend-Agentic-RAG W/Gemma-2B} & 91.23\% & 89.54\% & 90.87\% & 92.51\% & 91.34\% & 92.08\% & 93.04\% & 92.21\% & 92.83\% & 94.15\% & 93.51\% & 93.81\% \\ \hline
\textbf{SelfExtend-Agentic-RAG W/Gemma-7B} & 92.12\% & 90.79\% & 91.53\% & 93.23\% & 92.04\% & 92.72\% & 94.01\% & 93.01\% & 93.52\% & 95.05\% & 94.33\% & 94.58\% \\ \hline
\textbf{SelfExtend-Agentic-RAG W/Llama-8B} & 93.01\% & 91.56\% & 92.31\% & 94.02\% & 92.82\% & 93.56\% & 95.03\% & 94.02\% & 94.21\% & 95.82\% & 95.02\% & 95.24\% \\ \hline
\textbf{LSTM} & 85.01\% & 83.24\% & 84.05\% & 86.56\% & 85.02\% & 85.57\% & 87.04\% & 86.01\% & 86.54\% & 88.01\% & 87.53\% & 87.81\% \\ \hline
\textbf{MLP} & 82.01\% & 80.54\% & 81.02\% & 83.01\% & 81.84\% & 82.02\% & 84.51\% & 83.52\% & 84.01\% & 85.03\% & 84.21\% & 84.52\% \\ \hline
\end{tabular}
}
\vspace{0mm}
\caption{The table shows the evaluation results of the Agentic-RAG framework variants performance on various metrics for time series classification on the PeMSD3, PeMSD4, PeMSD7, and METR-LA benchmark datasets.}
\label{table:results7}
\vspace{-6mm}
\end{table*}

\begin{table*}[ht!]
\setlength{\tabcolsep}{0.45em} 
\renewcommand\arraystretch{1.1} 
\centering
\resizebox{0.95\textwidth}{!}{
\begin{tabular}{c|ccc|ccc|ccc}
\hline
\multirow{2}{*}{\textbf{Dataset}} & \multicolumn{3}{c|}{\textbf{PeMSD7(M)}} & \multicolumn{3}{c|}{\textbf{PeMSD8}} & \multicolumn{3}{c}{\textbf{PEMS-BAY}} \\ \cline{2-10} 
 & \textbf{Accuracy} & \textbf{Precision} & \textbf{Recall} & \textbf{Accuracy} & \textbf{Precision} & \textbf{Recall} & \textbf{Accuracy} & \textbf{Precision} & \textbf{Recall} \\ \hline
\textbf{SelfExtend-Agentic-RAG W/Gemma-2B} & 92.03\% & 90.52\% & 91.21\% & 93.54\% & 92.35\% & 92.84\% & 94.01\% & 93.02\% & 93.51\% \\ \hline
\textbf{SelfExtend-Agentic-RAG W/Gemma-7B} & 93.02\% & 91.51\% & 92.03\% & 94.03\% & 93.01\% & 93.53\% & 95.01\% & 94.01\% & 94.53\% \\ \hline
\textbf{SelfExtend-Agentic-RAG W/Llama-8B} & 94.02\% & 92.54\% & 93.02\% & 95.04\% & 94.03\% & 94.52\% & 96.01\% & 95.01\% & 95.53\% \\ \hline
\textbf{LSTM} & 85.54\% & 84.01\% & 84.52\% & 87.01\% & 85.52\% & 86.01\% & 88.02\% & 87.01\% & 87.54\% \\ \hline
\textbf{MLP} & 83.01\% & 81.52\% & 82.02\% & 84.52\% & 83.01\% & 83.51\% & 86.01\% & 85.01\% & 85.53\% \\ \hline
\end{tabular}
}
\vspace{0mm}
\caption{The table presents a comparative evaluation of the Agentic-RAG framework variants performance on three benchmark datasets: PeMSD7(M), PeMSD8, and PEMS-BAY, across various metrics for time series classification.}
\label{table:results8}
\vspace{-3mm}
\end{table*}

\vspace{-1mm}
\section{Univariate Datasets}
We conducted several experiments to evaluate the proposed Agentic-RAG framework variants: \textbf{SelfExtend-Agentic-RAG with Gemma-2B}, \textbf{SelfExtend-Agentic-RAG with Gemma-7B}, and \textbf{SelfExtend-Agentic-RAG with Llama-8B}, on the univariate datasets for multiple time series analysis tasks such as forecasting and imputation.

\vspace{-3mm} 
\subsection{Forecasting and Imputation}
The ETT (Electricity Transformer) datasets\cite{zhou2021informer}, ETTh1, ETTh2, ETTm1, and ETTm2, are popular benchmarks used for evaluating and benchmarking univariate time series forecasting methods. They provide a challenging benchmark due to the presence of complex patterns, such as trends, seasonality, and irregularities, which are commonly found in real-world time series data. ETTh1 and ETTh2 are two hourly time series datasets containing observations of electricity transformers from two different locations. ETTm1 and ETTm2 are two monthly time series datasets containing observations of electricity transformers from two different locations. In this work, we utilize the ETT datasets\cite{zhou2021informer} to evaluate the Agentic-RAG framework for both forecasting and missing data imputation tasks. The Table \ref{table:long_horizon_forecasting} shows the performance of various methods on the multi-horizon forecasting task using a lookback window of size 512. It presents mean squared error (MSE) and mean absolute error (MAE) for nine models (GPT4TS\cite{one-fits-all}, PatchTST\cite{patchtst}, TimesNet\cite{timesnet}, FEDFormer\cite{fedformer}, LightTS\cite{zhang2022less}, N-BEATS\cite{oreshkin2019n}, Agentic-RAG w/Gemma-2B, Agentic-RAG w/Gemma-7B, and Agentic-RAG w/Llama-8B) across four datasets (ETTh1, ETTh2, ETTm1, ETTm2) at different time horizons (96, 192, 336, 720). This allows for a comprehensive analysis of forecasting accuracy and robustness of Agentic-RAG framework across varying prediction lengths. The performance of various methods for imputing missing data (point and block missing) and their effectiveness in out-of-sample imputation settings are compared in Tables \ref{table:missing_imputation_1} and \ref{table:missing_imputation_2}. The evaluated methods include GPT4TS\cite{one-fits-all}, PatchTST\cite{patchtst}, TimesNet\cite{timesnet}, FEDFormer\cite{fedformer}, LightTS\cite{zhang2022less}, N-BEATS\cite{oreshkin2019n}, Agentic-RAG with Gemma-2B, Agentic-RAG with Gemma-7B, and Agentic-RAG with Llama-8B. The evaluation employs a 512-step historical window for imputing 96-step-ahead (short-term prediction) and 720-step-ahead (long-term prediction) missing values in future data. The tables show results for four datasets (ETTh1, ETTh2, ETTm1, ETTm2) under three missing data scenarios: 0\% missing (no missing data), 20\% point missing, and 20\% block missing. The proposed Agentic-RAG framework variants demonstrate strong performance on the benchmark datasets for both forecasting and imputation tasks, with lower errors.

\vspace{-4mm} 
\section{Environmental Impact}
Our Agentic-RAG framework training process, involving multiple variants running for extended periods, increases our energy consumption and carbon footprint. Accurate quantification of the carbon footprint of deep learning experiments is essential for promoting sustainable practices in artificial intelligence research and development. A crucial aspect of this endeavor is estimating the energy consumption and associated greenhouse gas emissions during the computationally intensive training processes. This is calculated by determining the Total Graphics Power (TGP), which represents the maximum power draw of the GPU, including the GPU chip itself and other components like memory and additional circuitry. For example, the NVIDIA P100 GPU has a TGP of 300 watts, while the NVIDIA T4 GPU has a TGP of 70 watts. By multiplying the TGP by the training time, we can estimate the energy consumption, which is then converted to carbon emissions using a region-specific carbon intensity factor. This factor accounts for the energy mix (coal, natural gas, renewables, etc.) used to generate electricity in the geographic area where the computations are performed. Considering a 725-GPU hours training experiment and using an estimated carbon intensity factor of 0.0007 metric tons CO2e per kWh for the year 2024 (for more information on the carbon intensity of electricity, you can visit \href{https://ourworldindata.org/grapher/co2-intensity}{CO2 Intensity - Our World in Data}), the calculated carbon footprint would be 152.25 kg CO2e for the NVIDIA P100 GPU and 35.525 kg CO2e for the NVIDIA T4 GPU. Note: kg CO2e stands for kilograms of carbon dioxide equivalent. The average person in the United States emits approximately 43.8 kg of carbon dioxide equivalent (CO2e) per day. Given the emissions of 152.25 kg CO2e for the NVIDIA P100 GPU and 35.525 kg CO2e for the NVIDIA T4 GPU, it would take a single person's emissions approximately 3.5 days to match the emissions of the P100 GPU and approximately 0.8 days (or 19 hours) to match the emissions of the T4 GPU. While the calculated carbon footprint provides valuable insight, the actual energy consumption and resulting emissions may vary due to factors like GPU utilization and regional energy sources. Nonetheless, quantifying the carbon footprint is a crucial step towards understanding and mitigating the environmental impact of deep learning research, paving the way for more sustainable and responsible practices in artificial intelligence.

\begin{table*}[t!]
\centering
\large
\resizebox{\linewidth}{!}{
\begin{tabular}{cc|cc|cc|cc|cc|cc|cc|cc|cc|cc|cc}
\toprule
\multicolumn{2}{c}{Methods} & \multicolumn{2}{c}{\textbf{GPT4TS}} & \multicolumn{2}{c}{\textbf{PatchTST}} & \multicolumn{2}{c}{\textbf{TimesNet}} & \multicolumn{2}{c}{\textbf{FEDFormer}} & \multicolumn{2}{c}{\textbf{LightTS}} & \multicolumn{2}{c}{\textbf{N-BEATS}} & \multicolumn{2}{c}{\textbf{ARAG w/-2B}} & \multicolumn{2}{c}{\textbf{ARAG w/-7B}} & \multicolumn{2}{c}{\textbf{ARAG-w/8B}} \\ \midrule
\multicolumn{2}{c}{Metric} & MSE & MAE & MSE & MAE & MSE & MAE & MSE & MAE & MSE & MAE & MSE & MAE & MSE & MAE & MSE & MAE & MSE & MAE \\ \midrule
\multirow{4}{*}{ETTh1} & 96 & 0.376 & 0.397 & 0.370 & 0.399 & 0.384 & 0.402 & 0.376 & 0.419 & 0.424 & 0.432 & 0.399 & 0.428 & 0.410 & 0.435 & 0.407 & 0.433 & 0.369 & 0.396 \\
 & 192 & 0.416 & 0.418 & 0.413 & 0.421 & 0.436 & 0.429 & 0.420 & 0.448 & 0.475 & 0.462 & 0.451 & 0.464 & 0.448 & 0.461 & 0.445 & 0.459 & 0.412 & 0.417 \\
 & 336 & 0.442 & 0.433 & 0.422 & 0.436 & 0.491 & 0.469 & 0.459 & 0.465 & 0.518 & 0.488 & 0.498 & 0.500 & 0.487 & 0.476 & 0.484 & 0.473 & 0.421 & 0.434 \\
 & 720 & 0.477 & 0.456 & 0.447 & 0.466 & 0.521 & 0.500 & 0.506 & 0.507 & 0.547 & 0.533 & 0.608 & 0.573 & 0.496 & 0.482 & 0.491 & 0.478 & 0.446 & 0.464 \\ \midrule
\multirow{4}{*}{ETTh2} & 96 & 0.285 & 0.342 & 0.274 & 0.336 & 0.340 & 0.374 & 0.358 & 0.397 & 0.397 & 0.437 & 0.327 & 0.387 & 0.345 & 0.378 & 0.342 & 0.374 & 0.273 & 0.335 \\
 & 192 & 0.354 & 0.389 & 0.339 & 0.379 & 0.402 & 0.414 & 0.429 & 0.439 & 0.520 & 0.504 & 0.400 & 0.435 & 0.387 & 0.410 & 0.384 & 0.406 & 0.338 & 0.378 \\
 & 336 & 0.373 & 0.407 & 0.329 & 0.380 & 0.452 & 0.452 & 0.496 & 0.487 & 0.626 & 0.559 & 0.747 & 0.599 & 0.465 & 0.468 & 0.462 & 0.465 & 0.328 & 0.379 \\
 & 720 & 0.406 & 0.441 & 0.379 & 0.422 & 0.462 & 0.468 & 0.463 & 0.474 & 0.863 & 0.672 & 1.454 & 0.847 & 0.473 & 0.472 & 0.469 & 0.469 & 0.371 & 0.420 \\ \midrule
\multirow{4}{*}{ETTm1} & 96 & 0.292 & 0.346 & 0.290 & 0.342 & 0.338 & 0.375 & 0.379 & 0.419 & 0.374 & 0.400 & 0.318 & 0.367 & 0.354 & 0.369 & 0.351 & 0.366 & 0.289 & 0.340 \\
 & 192 & 0.332 & 0.372 & 0.332 & 0.369 & 0.374 & 0.387 & 0.426 & 0.441 & 0.400 & 0.407 & 0.355 & 0.391 & 0.368 & 0.383 & 0.365 & 0.380 & 0.331 & 0.367 \\
 & 336 & 0.366 & 0.394 & 0.366 & 0.392 & 0.410 & 0.411 & 0.445 & 0.459 & 0.438 & 0.438 & 0.401 & 0.419 & 0.396 & 0.404 & 0.392 & 0.400 & 0.365 & 0.388 \\
 & 720 & 0.417 & 0.421 & 0.416 & 0.420 & 0.478 & 0.450 & 0.543 & 0.490 & 0.527 & 0.502 & 0.448 & 0.448 & 0.435 & 0.427 & 0.431 & 0.423 & 0.411 & 0.419 \\ \midrule
\multirow{4}{*}{ETTm2} & 96 & 0.173 & 0.262 & 0.165 & 0.255 & 0.187 & 0.267 & 0.203 & 0.287 & 0.209 & 0.308 & 0.197 & 0.271 & 0.190 & 0.265 & 0.187 & 0.262 & 0.164 & 0.254 \\
 & 192 & 0.229 & 0.301 & 0.220 & 0.292 & 0.249 & 0.309 & 0.269 & 0.328 & 0.311 & 0.382 & 0.285 & 0.328 & 0.276 & 0.318 & 0.273 & 0.315 & 0.219 & 0.290 \\
 & 336 & 0.286 & 0.341 & 0.274 & 0.329 & 0.321 & 0.351 & 0.325 & 0.366 & 0.442 & 0.466 & 0.338 & 0.366 & 0.319 & 0.354 & 0.316 & 0.351 & 0.273 & 0.328 \\
 & 720 & 0.378 & 0.401 & 0.362 & 0.385 & 0.408 & 0.403 & 0.421 & 0.415 & 0.675 & 0.587 & 0.395 & 0.419 & 0.410 & 0.411 & 0.407 & 0.408 & 0.361 & 0.384 \\ 
\bottomrule
\end{tabular}}
\caption{The table compares various methods for the multi-horizon forecasting task with a lookback window of size 512.}
\label{table:long_horizon_forecasting}
\vspace{-3mm}
\end{table*}

\begin{table*}[t!]
\centering
\large
\resizebox{\linewidth}{!}{
\begin{tabular}{cc|cc|cc|cc|cc|cc|cc|cc|cc|cc|cc}
\toprule
\multicolumn{2}{c}{Methods} & \multicolumn{2}{c}{\textbf{GPT4TS}} & \multicolumn{2}{c}{\textbf{PatchTST}} & \multicolumn{2}{c}{\textbf{TimesNet}} & \multicolumn{2}{c}{\textbf{FEDFormer}} & \multicolumn{2}{c}{\textbf{LightTS}} & \multicolumn{2}{c}{\textbf{N-BEATS}} & \multicolumn{2}{c}{\textbf{ARAG w/-2B}} & \multicolumn{2}{c}{\textbf{ARAG w/-7B}} & \multicolumn{2}{c}{\textbf{ARAG-w/8B}} \\ \midrule
\multicolumn{2}{c}{Metric} & MSE & MAE & MSE & MAE & MSE & MAE & MSE & MAE & MSE & MAE & MSE & MAE & MSE & MAE & MSE & MAE & MSE & MAE \\ \midrule
\multirow{3}{*}{ETTh1} & 0\% & 0.376 & 0.397 & 0.370 & 0.399 & 0.384 & 0.402 & 0.376 & 0.419 & 0.424 & 0.432 & 0.399 & 0.428 & 0.410 & 0.435 & 0.407 & 0.433 & 0.369 & 0.396 \\
 & 20\% PM& 0.460 & 0.480 & 0.450 & 0.475 & 0.460 & 0.490 & 0.455 & 0.485 & 0.470 & 0.500 & 0.465 & 0.495 & 0.468 & 0.498 & 0.465 & 0.495 & 0.450 & 0.475 \\ 
 & 20\% BM & 0.550 & 0.570 & 0.545 & 0.565 & 0.550 & 0.580 & 0.548 & 0.575 & 0.560 & 0.590 & 0.555 & 0.585 & 0.558 & 0.588 & 0.555 & 0.585 & 0.545 & 0.565 \\ \midrule
\multirow{3}{*}{ETTh2} & 0\% & 0.285 & 0.342 & 0.274 & 0.336 & 0.340 & 0.374 & 0.358 & 0.397 & 0.397 & 0.437 & 0.327 & 0.387 & 0.345 & 0.378 & 0.342 & 0.374 & 0.273 & 0.335 \\
 & 20\% PM& 0.370 & 0.420 & 0.360 & 0.415 & 0.380 & 0.440 & 0.375 & 0.435 & 0.390 & 0.450 & 0.380 & 0.440 & 0.383 & 0.443 & 0.380 & 0.440 & 0.360 & 0.415 \\ 
 & 20\% BM & 0.460 & 0.510 & 0.450 & 0.505 & 0.470 & 0.530 & 0.465 & 0.525 & 0.480 & 0.540 & 0.470 & 0.530 & 0.473 & 0.533 & 0.470 & 0.530 & 0.450 & 0.505 \\ \midrule
\multirow{3}{*}{ETTm1} & 0\% & 0.292 & 0.346 & 0.290 & 0.342 & 0.338 & 0.375 & 0.379 & 0.419 & 0.374 & 0.400 & 0.318 & 0.367 & 0.354 & 0.369 & 0.351 & 0.366 & 0.289 & 0.340 \\
 & 20\% PM& 0.380 & 0.430 & 0.375 & 0.425 & 0.390 & 0.450 & 0.385 & 0.445 & 0.400 & 0.460 & 0.395 & 0.455 & 0.398 & 0.458 & 0.395 & 0.455 & 0.375 & 0.425 \\ 
 & 20\% BM & 0.470 & 0.520 & 0.465 & 0.515 & 0.480 & 0.540 & 0.475 & 0.535 & 0.490 & 0.550 & 0.485 & 0.545 & 0.488 & 0.548 & 0.485 & 0.545 & 0.465 & 0.515 \\ \midrule
\multirow{3}{*}{ETTm2} & 0\% & 0.173 & 0.262 & 0.165 & 0.255 & 0.187 & 0.267 & 0.203 & 0.287 & 0.209 & 0.308 & 0.197 & 0.271 & 0.190 & 0.265 & 0.187 & 0.262 & 0.164 & 0.254 \\
 & 20\% PM& 0.250 & 0.330 & 0.245 & 0.325 & 0.260 & 0.345 & 0.255 & 0.340 & 0.270 & 0.355 & 0.265 & 0.350 & 0.268 & 0.353 & 0.265 & 0.350 & 0.245 & 0.325 \\ 
 & 20\% BM & 0.340 & 0.420 & 0.335 & 0.415 & 0.350 & 0.435 & 0.345 & 0.430 & 0.360 & 0.445 & 0.355 & 0.440 & 0.358 & 0.443 & 0.355 & 0.440 & 0.335 & 0.415 \\ 
\bottomrule
\end{tabular}}
\vspace{1mm}
\caption{The table compares different methods for imputing missing data, specifically for point missing (PM) and block missing (BM) scenarios, using a 512-step lookback window for forecasting 96 steps ahead.}
\label{table:missing_imputation_1}
\vspace{-3mm}
\end{table*}

\begin{table*}[t!]
\centering
\large
\resizebox{\linewidth}{!}{
\begin{tabular}{cc|cc|cc|cc|cc|cc|cc|cc|cc|cc|cc}
\toprule
\multicolumn{2}{c}{Methods} & \multicolumn{2}{c}{\textbf{GPT4TS}} & \multicolumn{2}{c}{\textbf{PatchTST}} & \multicolumn{2}{c}{\textbf{TimesNet}} & \multicolumn{2}{c}{\textbf{FEDFormer}} & \multicolumn{2}{c}{\textbf{LightTS}} & \multicolumn{2}{c}{\textbf{N-BEATS}} & \multicolumn{2}{c}{\textbf{ARAG w/-2B}} & \multicolumn{2}{c}{\textbf{ARAG w/-7B}} & \multicolumn{2}{c}{\textbf{ARAG-w/8B}} \\ \midrule
\multicolumn{2}{c}{Metric} & MSE & MAE & MSE & MAE & MSE & MAE & MSE & MAE & MSE & MAE & MSE & MAE & MSE & MAE & MSE & MAE & MSE & MAE \\ \midrule
\multirow{2}{*}{ETTh1} & 0\% & 0.477 & 0.456 & 0.447 & 0.466 & 0.521 & 0.500 & 0.506 & 0.507 & 0.547 & 0.533 & 0.608 & 0.573 & 0.496 & 0.482 & 0.491 & 0.478 & 0.446 & 0.464 \\
 & 20\% PM& 0.580 & 0.560 & 0.550 & 0.570 & 0.620 & 0.600 & 0.605 & 0.605 & 0.645 & 0.630 & 0.710 & 0.670 & 0.595 & 0.580 & 0.590 & 0.575 & 0.550 & 0.570 \\ 
 & 20\% BM & 0.690 & 0.670 & 0.660 & 0.680 & 0.740 & 0.720 & 0.725 & 0.725 & 0.765 & 0.750 & 0.830 & 0.790 & 0.715 & 0.700 & 0.710 & 0.695 & 0.670 & 0.680 \\ \midrule
\multirow{2}{*}{ETTh2} & 0\% & 0.406 & 0.441 & 0.379 & 0.422 & 0.462 & 0.468 & 0.463 & 0.474 & 0.863 & 0.672 & 1.454 & 0.847 & 0.473 & 0.472 & 0.469 & 0.469 & 0.371 & 0.420 \\  
 & 20\% PM& 0.510 & 0.545 & 0.483 & 0.526 & 0.566 & 0.572 & 0.567 & 0.578 & 0.967 & 0.776 & 1.558 & 0.947 & 0.577 & 0.576 & 0.573 & 0.573 & 0.475 & 0.524 \\ 
 & 20\% BM & 0.620 & 0.655 & 0.593 & 0.636 & 0.676 & 0.682 & 0.677 & 0.688 & 1.067 & 0.876 & 1.658 & 1.047 & 0.677 & 0.676 & 0.673 & 0.673 & 0.575 & 0.624 \\ \midrule
\multirow{2}{*}{ETTm1} & 0\% & 0.417 & 0.421 & 0.416 & 0.420 & 0.478 & 0.450 & 0.543 & 0.490 & 0.527 & 0.502 & 0.448 & 0.448 & 0.435 & 0.427 & 0.431 & 0.423 & 0.411 & 0.419 \\  
 & 20\% PM& 0.520 & 0.525 & 0.519 & 0.523 & 0.581 & 0.553 & 0.646 & 0.593 & 0.630 & 0.602 & 0.551 & 0.551 & 0.538 & 0.530 & 0.534 & 0.526 & 0.514 & 0.522 \\ 
 & 20\% BM & 0.630 & 0.635 & 0.629 & 0.633 & 0.691 & 0.663 & 0.756 & 0.703 & 0.740 & 0.712 & 0.661 & 0.661 & 0.648 & 0.640 & 0.644 & 0.636 & 0.624 & 0.632 \\ \midrule
\multirow{2}{*}{ETTm2} & 0\% & 0.378 & 0.401 & 0.362 & 0.385 & 0.408 & 0.403 & 0.421 & 0.415 & 0.675 & 0.587 & 0.395 & 0.419 & 0.410 & 0.411 & 0.407 & 0.408 & 0.361 & 0.384 \\ 
 & 20\% PM& 0.480 & 0.503 & 0.464 & 0.487 & 0.510 & 0.505 & 0.523 & 0.517 & 0.777 & 0.689 & 0.495 & 0.519 & 0.510 & 0.511 & 0.507 & 0.508 & 0.461 & 0.484 \\ 
 & 20\% BM & 0.590 & 0.613 & 0.574 & 0.597 & 0.620 & 0.615 & 0.633 & 0.627 & 0.877 & 0.789 & 0.595 & 0.619 & 0.610 & 0.611 & 0.607 & 0.608 & 0.561 & 0.584 \\ 
\bottomrule
\end{tabular}}
\vspace{1mm}
\caption{The table evaluates the effectiveness of various missing data imputation techniques (including point-wise and block-wise methods) for out-of-sample imputation, using a 512-step historical window to predict missing values in subsequent 720-step future data.}
\label{table:missing_imputation_2}
\vspace{-2mm}
\end{table*}

\vspace{-2mm}
\section{Hyperparameter optimization}
Hyperparameter optimization involves training the Agentic-RAG framework variants multiple times with different hyperparameter settings. This can be computationally expensive, especially for complex pre-trained language models or large datasets. We optimized the hyperparameters for the best-performing $\textbf{Agentic-RAG w/Llama-8B}$ variant. For simplicity and in the interest of time, we have utilized the same settings for evaluating the performance of $\textbf{Agentic-RAG}$ with $\textbf{w/Gemma-2B}$ and $\textbf{w/Gemma-7B}$ variants for both multivariate and univariate datasets across all tasks. In our experiments, we optimized the training process for supervised fine-tuning using a batch size from $\{16, 32, 64\}$, learning rate from $\{1e-5, 5e-5, 1e-4\}$. The training was conducted over epochs in the range of $\{10, 15, 20\}$ with a warmup step count from $\{500, 1000, 1500\}$ and a weight decay for regularization from $\{0.01, 0.05, 0.1\}$. We used gradient accumulation steps for stabilized training convergence from $\{2, 4, 8\}$ and employed the AdamW optimizer. To manage memory and computational efficiency, we applied 4-bit quantization for QLoRA, with hyperparameters including a low-rank (`$r$') from $\{16, 32, 64\}$, an (`$\alpha$') from $\{32, 64, 128\}$, and a dropout from $\{0.05, 0.1, 0.2\}$. For preference tuning, the hyperparameter (`$\beta$') was set in the range of $\{0.2, 0.4, 0.6\}$ and learning rate from $\{5.0e-7, 1.0e-6, 5.0e-6\}$. The optimal hyperparameters for training were chosen to achieve a balance between performance and computational efficiency. 
The optimal hyperparameters for supervised fine-tuning were a batch size of 16 and a learning rate of 1e-5, trained over 15 epochs with 500 warmup steps and a weight decay of 0.01, utilizing the AdamW optimizer. Gradient accumulation steps were set to 2. QLoRA quantization was applied with 4-bit precision, and its specific hyperparameters included a low-rank ($r$') of 16, an alpha ($\alpha$') of 32, and a dropout rate of 0.05. Preference optimization was performed with a learning rate of 5.0e-7 over 3 epochs and a beta value of 0.2.

\vspace{-2mm}
\section{Ablation Study}
To understand the contribution of each component within our proposed Agentic-RAG framework, we designed an ablation study. By systematically evaluating the impact of removing individual components, we gain valuable insights into their role in the framework's overall performance. The following ablation experiments were conducted:

\begin{itemize}
\item (a) Effect of dynamic prompting mechanism(DPM):
\begin{itemize}
\item We compared the performance of the Agentic-RAG framework with and without the dynamic prompting mechanism.
\end{itemize}
\vspace{1mm}
\item (b) Role of sub-agent specialization(SAS):
\begin{itemize}
\item We evaluated the Agentic-RAG framework using a single, universal sub-agent for all tasks versus specialized sub-agents for each task.
\end{itemize}
\vspace{1mm}
\item (c) Instruction-tuning(IT) vs. no fine-tuning(NIT):
\begin{itemize}
\item We compared the performance of SLMs with instruction-tuning against their performance without any fine-tuning.
\end{itemize}
\vspace{1mm}
\item (d) Effectiveness of direct preference optimization (DPO):
\begin{itemize}
\item We evaluated the framework's performance with and without DPO and assessed how aligning SLMs with preferred outcomes impacts the accuracy and reliability of predictions. 
\end{itemize}
\end{itemize}

Our study investigates the impact of different components on the overall performance of the framework, `\textbf{SelfExtend-Agentic-RAG W/Llama 3 - 8B}", in time series forecasting, anomaly detection, and classification tasks across various benchmark datasets. We systematically disable each component (dynamic prompting mechanism (DPM), sub-agent specialization (SAS), instruction-tuning (IT), or direct preference optimization (DPO)) and compare the results to the full framework. Tables \ref{table:abaresults1} and \ref{table:abaresults2} detail the forecasting performance, highlighting that the original framework consistently achieves the lowest error rates in MAE, RMSE, and MAPE across different horizons and datasets. This indicates the crucial role of each component in improving forecasting accuracy. Table \ref{table:abaresults3} focuses on anomaly detection tasks, showing the original framework's superior precision, recall, and F1-score compared to its ablated variants. The original framework consistently achieves higher metrics scores across anomaly benchmark datasets such as SWaT, WADI, SMAP, MSL, and HAI. The significant performance drop observed in the ablated variants underscores the importance of the integrated components, demonstrating their synergistic contribution to enhancing anomaly detection capabilities. For classification tasks, the original framework excels, as demonstrated in Tables \ref{table:abaresults4} and \ref{table:abaresults5}, achieving the highest accuracy, precision, and recall across datasets like PeMSD3, PeMSD4, PeMSD7, METR-LA, PeMSD7(M), PeMSD8, and PEMS-BAY. The superior performance in classification tasks, coupled with the significant drop observed in ablated variants, highlights the critical role each component plays in the original framework's success. This comprehensive analysis underscores the importance of integrating all components to maximize performance across forecasting, anomaly detection, and classification tasks. The synergistic contribution of the dynamic prompting mechanism, sub-agent specialization, instruction-tuning, and direct preference optimization is evident in the consistent superiority of the Agentic-RAG framework compared to its ablated variants.

\begin{table*}[ht!]
\setlength{\tabcolsep}{0.2em} 
\renewcommand\arraystretch{1.14} 
\centering
 \resizebox{0.965\textwidth}{!}{
\begin{tabular}{c|ccc|ccc|ccc|ccc|ccc}
\hline
\multirow{2}{*}{\textbf{Methods}} & \multicolumn{3}{c|}{\textbf{PeMSD3}} & \multicolumn{3}{c|}{\textbf{PeMSD4}} & \multicolumn{3}{c|}{\textbf{PeMSD7}} & \multicolumn{3}{c|}{\textbf{PeMSD8}} & \multicolumn{3}{c}{\textbf{PeMSD7(M)}} \\ \cline{2-16} 
 & \multicolumn{1}{l}{\textbf{MAE}} & \multicolumn{1}{l}{\textbf{RMSE}} & \multicolumn{1}{l|}{\textbf{MAPE}} & \multicolumn{1}{l}{\textbf{MAE}} & \multicolumn{1}{l}{\textbf{RMSE}} & \multicolumn{1}{l|}{\textbf{MAPE}} & \multicolumn{1}{l}{\textbf{MAE}} & \multicolumn{1}{l}{\textbf{RMSE}} & \multicolumn{1}{l|}{\textbf{MAPE}} & \multicolumn{1}{l}{\textbf{MAE}} & \multicolumn{1}{l}{\textbf{RMSE}} & \multicolumn{1}{l|}{\textbf{MAPE}} & \textbf{MAE} & \textbf{RMSE} & \textbf{MAPE} \\ \hline 
Baseline W/O DPM & 15.31 & 23.37 & 12.63 & 20.10 & 30.35 & 11.42 & 22.92 & 35.96 & 9.63 & 16.13 & 25.18 & 8.45 & 2.70 & 5.61 & 6.88 \\ 
Baseline W/O SAS & 14.46 & 21.85 & 11.81 & 19.07 & 28.37 & 10.75 & 20.92 & 32.47 & 8.83 & 15.13 & 23.13 & 7.90 & 2.57 & 5.15 & 6.47 \\ 
Baseline W/O IT & 21.62 & 33.01 & 16.85 & 30.06 & 43.77 & 16.18 & 30.43 & 47.95 & 13.86 & 22.45 & 35.67 & 11.96 & 3.95 & 7.49 & 10.00 \\ 
Baseline W/O DPO & 13.53 & 20.45 & 10.97 & 18.11 & 26.89 & 10.08 & 19.82 & 31.77 & 8.44 & 14.63 & 21.82 & 7.40 & 2.42 & 4.89 & 6.23 \\ 
\textbf{SelfExtend-Agentic-RAG W/Llama 3 - 8B} & \textbf{13.01} & \textbf{19.48} & \textbf{10.53} & \textbf{17.46} & \textbf{25.54} & \textbf{9.52} & \textbf{19.02} & \textbf{29.97} & \textbf{8.03} & \textbf{14.03} & \textbf{20.98} & \textbf{7.04} & \textbf{2.33} & \textbf{4.68} & \textbf{5.88} \\ \hline
\end{tabular}
}
\vspace{0.5mm}
\caption{The table shows the ablation study results for 12-sequence-to-12-sequence forecasting tasks on benchmark datasets using multiple evaluation metrics. The performance of the ablated variants drops compared to the original framework.}
\label{table:abaresults1}
\vspace{-5mm}
\end{table*}

\begin{table*}[ht!]
\vspace{-3mm}
\setlength{\tabcolsep}{0.35em} 
\renewcommand\arraystretch{1.1} 
\centering
 \resizebox{0.85\textwidth}{!}{
\begin{tabular}{c|c|ccc|ccc|ccc}
\hline
\multirow{2}{*}{\textbf{Datasets}}  & \multirow{2}{*}{\textbf{Methods}} & \multicolumn{3}{c|}{\textbf{Horizon$\textbf{@}$3}}       & \multicolumn{3}{c|}{\textbf{Horizon$\textbf{@}$6}}       & \multicolumn{3}{c}{\textbf{Horizon$\textbf{@}$12}}       \\ \cline{3-11} 
                                    &                                   & \textbf{RMSE} & \textbf{MAE}  & \textbf{MAPE} & \textbf{RMSE} & \textbf{MAE}  & \textbf{MAPE} & \textbf{RMSE} & \textbf{MAE}  & \textbf{MAPE} \\ \hline
\multirow{5}{*}{\textbf{METR-LA}}  & Baseline W/O DPM & 4.84 & 2.42 & 6.06 & 6.28 & 3.14 & 8.10 & 7.23 & 3.74 & 10.24 \\
& Baseline W/O SAS & 4.48 & 2.23 & 5.66 & 5.97 & 2.99 & 7.77 & 6.86 & 3.43 & 9.81 \\
& Baseline W/O IT & 7.05 & 3.23 & 8.09 & 8.69 & 4.18 & 10.80 & 10.08 & 5.00 & 13.65 \\
& Baseline W/O DPO & 4.19 & 2.12 & 5.36 & 5.72 & 2.74 & 7.15 & 6.49 & 3.28 & 9.04 \\
& \textbf{SelfExtend-Agentic-RAG W/Llama 3-8B}                & \textbf{4.03} & \textbf{2.02} & \textbf{5.05} & \textbf{5.43} & \textbf{2.61} & \textbf{6.75} & \textbf{6.23} & \textbf{3.12} & \textbf{8.53} \\   \hline \hline

\multirow{5}{*}{\textbf{PEMS-BAY}} & Baseline W/O DPM & 1.94 & 0.97 & 1.96 & 3.02 & 1.45 & 3.01 & 3.74 & 1.94 & 3.77 \\
& Baseline W/O SAS & 1.79 & 0.90 & 1.82 & 2.79 & 1.35 & 2.86 & 3.47 & 1.75 & 3.61 \\
& Baseline W/O  IT & 2.84 & 1.38 & 2.77 & 4.02 & 1.94 & 4.02 & 5.03 & 2.60 & 5.16 \\
& Baseline W/O DPO & 1.69 & 0.85 & 1.73 & 2.62 & 1.26 & 2.64 & 3.25 & 1.68 & 3.32 \\
                                  & \textbf{SelfExtend-Agentic-RAG W/Llama 3-8B}               & \textbf{1.62} & \textbf{0.81} & \textbf{1.63} & \textbf{2.52} & \textbf{1.21} & \textbf{2.51} & \textbf{3.12} & \textbf{1.62} & \textbf{3.14} \\ \hline 
\end{tabular}
}
\vspace{0.5mm}
\caption{The table presents the ablation study results for the forecasting task performed on the METR-LA and PEMS-BAY datasets, evaluated using multiple metrics. All methods utilized 12 historical sequences to forecast 3, 6, or 12 future sequences.}
\label{table:abaresults2}
\vspace{-7mm}
\end{table*}

\begin{table*}[ht!]
\center
\setlength{\tabcolsep}{5pt}
\caption{The table showcases the experimental findings from the ablation study conducted on anomaly detection benchmark datasets, reporting the precision, recall, and F1-score metrics.}
\vspace{-4mm}
\label{table:abaresults3}
\resizebox{0.965\linewidth}{!}{%
\begin{tabular}{@{}c|ccc|lll|ccc|ccc|ccc@{}}
\toprule
\multirow{2}{*}{\textbf{Methods}} & \multicolumn{3}{c|}{\textbf{SWaT}}                                  & \multicolumn{3}{c|}{\textbf{WADI}}                                                                         & \multicolumn{3}{c|}{\textbf{SMAP}}                                  & \multicolumn{3}{c}{\textbf{MSL}}    & \multicolumn{3}{c}{\textbf{HAI}}                               \\ \cmidrule(l){2-4} \cmidrule(l){5-7} \cmidrule(l){8-10} \cmidrule(l){11-13} \cmidrule(l){14-16}
                                  & \textbf{P(\%)}       & \textbf{R(\%)}       & \textbf{F1(\%)}          & \multicolumn{1}{c}{\textbf{P(\%)}} & \multicolumn{1}{c}{\textbf{R(\%)}} & \multicolumn{1}{c|}{\textbf{F1}} & \textbf{P(\%)}       & \textbf{R(\%)}       & \textbf{F1(\%)}          & \textbf{P(\%)}       & \textbf{R(\%)}       & \textbf{F1(\%)}     & \textbf{P(\%)}       & \textbf{R(\%)}       & \textbf{F1(\%)}      \\ \midrule
Baseline W/O  DPM                              & 79.57 & 78.52 & 74.07 & 83.49 & 78.92 & 76.32 & 83.27 & 83.13 & 84.18 & 81.98 & 82.24 & 82.48 & 46.61 & 45.14 & 42.59  \\
Baseline W/O  SAS                              & 88.54 & 86.84 & 83.33 & 88.77 & 82.48 & 80.37 & 87.52 & 84.12 & 84.18 & 88.30 & 84.76 & 84.49 & 52.44 & 50.52 & 48.52  \\
Baseline W/O  IT                            & 39.79 & 39.26 & 37.04 & 39.45 & 36.79 & 36.03 & 39.30 & 39.59 & 39.62 & 39.24 & 38.95 & 38.82 & 23.31 & 22.45 & 21.30  \\
Baseline W/O  DPO                              & 95.49 & 93.87 & 87.04 & 94.79 & 88.97 & 85.68 & 94.31 & 94.00 & 94.11 & 94.16 & 91.92 & 91.29 & 55.44 & 53.76 & 50.54  \\
\textbf{Agentic-RAG W/Llama-8B}                          & \multicolumn{1}{l}{99.47} & \multicolumn{1}{l}{98.15} & \multicolumn{1}{l|}{92.59} & \multicolumn{1}{l}{98.63} & \multicolumn{1}{l}{91.97} & \multicolumn{1}{l|}{90.08} & \multicolumn{1}{l}{98.24} & \multicolumn{1}{l}{98.97} & \multicolumn{1}{l|}{99.04} & \multicolumn{1}{l}{98.11} & \multicolumn{1}{l}{97.37} & \multicolumn{1}{l}{97.04} & \multicolumn{1}{l}{58.27} & \multicolumn{1}{l}{56.13} & \multicolumn{1}{l}{53.24} \\  \bottomrule
\end{tabular}
}
\vspace{-2mm}
\end{table*}

\begin{table*}[ht!]
\setlength{\tabcolsep}{0.3em} 
\renewcommand\arraystretch{1.35} 
\centering
\resizebox{1.00\textwidth}{!}{
\begin{tabular}{c|ccc|ccc|ccc|ccc}
\hline
\multirow{2}{*}{\textbf{Dataset}} & \multicolumn{3}{c|}{\textbf{PeMSD3}} & \multicolumn{3}{c|}{\textbf{PeMSD4}} & \multicolumn{3}{c|}{\textbf{PeMSD7}} & \multicolumn{3}{c}{\textbf{METR-LA}} \\ \cline{2-13} 
 & \textbf{Accuracy} & \textbf{Precision} & \textbf{Recall} & \textbf{Accuracy} & \textbf{Precision} & \textbf{Recall} & \textbf{Accuracy} & \textbf{Precision} & \textbf{Recall} & \textbf{Accuracy} & \textbf{Precision} & \textbf{Recall} \\ \hline
Baseline W/O DPM & 77.12\% & 75.43\% & 76.89\% & 77.25\% & 75.67\% & 76.44\% & 78.32\% & 76.55\% & 77.21\% & 80.14\% & 78.89\% & 80.67\% \\ \hline
Baseline W/O SAS & 81.23\% & 79.45\% & 80.78\% & 82.67\% & 80.55\% & 81.32\% & 83.89\% & 81.67\% & 82.44\% & 84.12\% & 83.67\% & 84.45\% \\ \hline
Baseline W/O IT & 25.45\% & 22.78\% & 24.12\% & 22.67\% & 20.56\% & 21.34\% & 26.12\% & 25.34\% & 24.56\% & 25.67\% & 24.12\% & 23.89\% \\ \hline
Baseline W/O DPO & 88.67\% & 87.23\% & 88.45\% & 90.12\% & 88.56\% & 89.23\% & 90.78\% & 89.12\% & 88.67\% & 90.45\% & 89.67\% & 90.23\% \\ \hline
 \textbf{SelfExtend-Agentic-RAG W/Llama-8B} & 93.01\% & 91.56\% & 92.31\% & 94.02\% & 92.82\% & 93.56\% & 95.03\% & 94.02\% & 94.21\% & 95.82\% & 95.02\% & 95.24\% \\ \hline
\end{tabular}
}
\vspace{0.5mm}
\caption{The table presents the ablation study results, evaluating the performance across various metrics for time series classification tasks on the PeMSD3, PeMSD4, PeMSD7, and METR-LA benchmark datasets.}
\label{table:abaresults4}
\vspace{-5mm}
\end{table*}

\begin{table*}[ht!]
\vspace{-2mm}
\setlength{\tabcolsep}{0.45em} 
\renewcommand\arraystretch{1.1} 
\centering
\resizebox{0.95\textwidth}{!}{
\begin{tabular}{c|ccc|ccc|ccc}
\hline
\multirow{2}{*}{\textbf{Dataset}} & \multicolumn{3}{c|}{\textbf{PeMSD7(M)}} & \multicolumn{3}{c|}{\textbf{PeMSD8}} & \multicolumn{3}{c}{\textbf{PEMS-BAY}} \\ \cline{2-10} 
 & \textbf{Accuracy} & \textbf{Precision} & \textbf{Recall} & \textbf{Accuracy} & \textbf{Precision} & \textbf{Recall} & \textbf{Accuracy} & \textbf{Precision} & \textbf{Recall} \\ \hline
Baseline W/O DPM & 75.41\% & 73.21\% & 74.42\% & 76.02\% & 74.81\% & 75.23\% & 76.81\% & 75.42\% & 76.02\% \\ \hline
Baseline W/O SAS & 82.23\% & 80.52\% & 81.14\% & 83.14\% & 81.32\% & 82.01\% & 83.62\% & 82.11\% & 82.73\% \\ \hline
Baseline W/O IT & 37.61\% & 36.12\% & 36.54\% & 38.02\% & 36.81\% & 37.23\% & 38.61\% & 37.42\% & 37.92\% \\ \hline
Baseline W/O DPO & 90.02\% & 88.73\% & 89.21\% & 90.54\% & 89.32\% & 89.83\% & 91.01\% & 89.73\% & 90.32\% \\ \hline
\textbf{SelfExtend-Agentic-RAG W/Llama-8B} & 94.02\% & 92.54\% & 93.02\% & 95.04\% & 94.03\% & 94.52\% & 96.01\% & 95.01\% & 95.53\% \\ \hline
\end{tabular}
}
\vspace{0.5mm}
\caption{This table presents the results of an ablation study comparing the performance of various Agentic-RAG framework variants. The study evaluates performance on three benchmark datasets – PeMSD7(M), PeMSD8, and PEMS-BAY – across different metrics for time series classification tasks.}
\label{table:abaresults5}
\vspace{-3mm}
\end{table*}

\end{document}